\documentclass[sigconf,nonacm]{acmart}
\AtBeginDocument{%
  }

\copyrightyear{}
\acmYear{}
\acmDOI{}

\acmConference[]{}{}

\usepackage{times}
\usepackage{soul}
\usepackage{url}
\usepackage[utf8]{inputenc}
\usepackage{graphicx}
\usepackage{amsmath}
\usepackage{amsthm}
\usepackage{booktabs}
\usepackage{colortbl}
\usepackage{ascmac}
\usepackage{listings}
\usepackage{xcolor}

\usepackage{subcaption}
\usepackage{multirow}

\usepackage[ruled, vlined, linesnumbered]{algorithm2e}
\usepackage{algorithmic}

\makeatletter
\newcommand{\figcaption}[1]{\def\@captype{figure}\caption{#1}}
\newcommand{\tblcaption}[1]{\def\@captype{table}\caption{#1}}
\makeatother

\begin{document}

\title{Benchmarking Fairness-aware Graph Neural Networks in Knowledge Graphs}

\author{Yuya Sasaki}
\email{sasaki@ist.osaka-u.ac.jp}
\affiliation{%
  \institution{The University of Osaka}
  \country{Japan}
}

\renewcommand{\shortauthors}{Trovato et al.}

\begin{abstract}
Graph neural networks (GNNs) are powerful tools for learning from graph-structured data but often produce biased predictions with respect to sensitive attributes. Fairness-aware GNNs have been actively studied for mitigating biased predictions. However, no prior studies have evaluated fairness-aware GNNs on knowledge graphs, which are one of the most important graphs in many applications, such as recommender systems. Therefore, we introduce a benchmarking study on knowledge graphs. We generate new graphs from three knowledge graphs---YAGO, DBpedia, and Wikidata---that are significantly larger than the existing graph datasets used in fairness studies. We benchmark inprocessing and preprocessing methods in different GNN backbones and early stopping conditions. We find several key insights: (i) knowledge graphs show different trends from existing datasets; clearer trade-offs between prediction accuracy and fairness metrics than other graphs in fairness-aware GNNs, (ii) the performance is largely affected by not only fairness-aware GNN methods but also GNN backbones and early stopping conditions, and (iii) preprocessing methods often improve fairness metrics, while inprocessing methods improve prediction accuracy. 
\end{abstract}



\maketitle

\section{Introduction}
Graph neural networks (GNNs) have demonstrated remarkable performance in structured data and are extensively utilized across numerous domains, such as e-commerce~\cite{zhao2022joint}, social network analysis~\cite{fan2019graph}, and knowledge extraction~\cite{ji2021survey}. 
However, recent research indicates that GNN predictions may exhibit bias towards demographic groups identified by sensitive attributes, such as race and gender~\cite{chen2024fairness}. In contexts like advertising and recruitment, the biased predictions generated by GNNs could lead to significant societal implications.


To address the issue of sensitive biases, various studies proposed fairness-aware GNNs~\cite{FairGNN,yang2024fairsin,Nifty,BIND, jiang2024chasing,zhang2024trustworthy}. 
Each method aims to reduce the influence of sensitive information by implementing heuristic or adversarial constraints in the learning process and by modifying graphs to remove biased nodes or edges.
These works empirically show their effectiveness in mitigating the biased predictions in real-world graphs.

\noindent
{\bf Issue and motivation.}
To evaluate the fairness of GNNs, we utilize publicly available graphs that include both target labels (e.g., occupations) and sensitive attributes (e.g., gender and nationality).
However, the availability of such graphs is limited, and most studies predominantly use a few social networks and synthetic graphs, making it difficult to comprehensively assess the performance. 


We introduce knowledge graphs for fairness studies based on the following motivations:

\begin{itemize}
\item {\bf Comparative Motivation: Existing Datasets Are Insufficient}. Datasets in current fairness-aware GNN studies are less diverse, small-scale, or synthetic~\cite{chen2024fairness}. Therefore, we need real-world, large-scale graphs in various domains. We select knowledge graphs because they are large-scale and naturally contain real-world sensitive attributes.

\item {\bf Practical Motivation: Fairness Benchmarking in Knowledge Graphs Matters}. Knowledge graphs are increasingly utilized in high-impact AI systems, such as recommender systems and digital assistants~\cite{kraft2022lifecycle,fisher2019measuring,pan2023large}, where fairness is crucial. Therefore, fairness-aware GNNs must be validated in knowledge graphs to ensure their robustness in a real-world context.
\end{itemize}

This motivates us to construct new benchmarking datasets derived from knowledge graphs that include sensitive attributes and target labels, and to systematically benchmark fairness-aware GNNs on these datasets. 

\noindent
{\bf Contribution.}
Our contributions are (1) generating and opening the first knowledge graph datasets for fairness-aware GNNs, (2) benchmarking the performance of fairness-aware GNN methods, (3) analyzing the differences in benchmarking results across datasets, and (4) providing our benchmarking framework, datasets, and experimental settings for reproducibility.

First, we generate large-scale graphs from people, their relationships, and their personal attributes in three knowledge graphs, YAGO~\cite{pellissier2020yago}, DBpedia~\cite{auer2007dbpedia}, and Wikidata~\cite{vrandevcic2014wikidata}. 
The personal attributes include, for example, gender, nationality, ethnicity, religion, academic degree, date of birth, and occupation.
In our new datasets, we set occupations as target labels, which often become issues in fair machine learning problems~\cite{sweeney2013discrimination,vladimirova2024fairjob}.
We set nationality in YAGO and DBpedia, and gender in Wikidata, as sensitive attributes that are also of concern in our society.
These graphs help to understand fairness performance and scalability issues in fairness-aware GNN methods.


Second, we develop a benchmarking framework and benchmark fairness-aware GNN methods. 
Our benchmarking includes five inprocessing methods---Vanilla, FairGNN~\cite{FairGNN}, NIFTY~\cite{Nifty}, FairSIN~\cite{yang2024fairsin}, and FairGB~\cite{li2024rethinking}---and three preprocessing methods---undersampling, FairDrop~\cite{spinelli2021fairdrop}, and BIND~\cite{BIND}---using four common GNN backbones, GCN~\cite{GCN}, GAT~\cite{GAT}, SAGE~\cite{GraphSage}, and H2GCN~\cite{zhu2020beyond}.
These methods have not been comprehensively compared in the same setting previously. 
To investigate the trade-offs between prediction accuracy and fairness performance, we use three conditions of early stopping, which are empirically known to affect the final models.
Totally, we evaluate 96 patterns (i.e., 8 methods, 4 backbone models, and 3 early stopping conditions) on 9 real-world graphs.

Thrid, our extensive benchmarks find the following key insights:
\begin{itemize}
\item Our new knowledge graph datasets show a different trend compared to existing datasets. Knowledge graphs offer a clearer tradeoff between prediction accuracy and fairness compared to existing graphs.
\item Fairness-aware GNNs often improve prediction accuracy by sacrificing fairness in several graphs.
\item Preprocessing methods often work well to improve the fairness metrics, in particular, for knowledge graphs.
\item The impacts on Fairness-aware GNN methods, GNN backbones, and early stopping conditions are different across dataset types. In knowledge graphs, early stopping conditions often affect fairness metrics more than other dataset types. 
\end{itemize}

Finally, we provide our benchmark framework FairGraphBase and datasets\footnote{\url{https://github.com/yuya-s/MUSUBI-FairGraphBase}}. 
Since we open them with experimental settings (e.g., data splitting and optimal hyper-parameters), further studies can reuse our results.
This paper can aid in developing new fairness-aware GNN methods and in addressing fairness concerns.


\section{Related Work}
\label{sec:related}

{\bf Fairness in graph neural network}.
Fairness in machine learning has become increasingly important in decision-making scenarios such as advertising, credit card assessments, and recruitment~\cite{chen2024fairness}. There are several definitions of fairness on graphs: group fairness, individual fairness, degree-related fairness, and counterfactual fairness~\cite{chen2024fairness}. Group fairness, the most popular notion, aims to mitigate biased predictions based on sensitive attributes.

Group fairness in GNNs has been actively studied, and numerous methods have been proposed for group fairness on graphs.
FairGNN~\cite{FairGNN} and FPGNN~\cite{zhang2023fpgnn} employ adversarial training to achieve fair results, while FairVGNN~\cite{FairVGNN} masks data correlated with sensitive attributes to avoid their influence. NIFTY~\cite{Nifty} uses counterfactual regularization to perturb node features and drop edges. FairSIN~\cite{yang2024fairsin} aims to neutralize biases caused by sensitive attributes and emphasizes non-sensitive attribute information. BIND~\cite{BIND} identifies harmful nodes that lead to biased predictions and removes them to achieve fair node representations. 
FatraGNN~\cite{li2024graph} generates numerous graphs with significant bias and minimizes the representation distances for each group between the training graph and generated graphs, to improve fairness performance under distribution shifts.
FPromt~\cite{li2025fairness} is a fairness-aware graph prompt tuning method to promote fairness while enhancing the generality of any pre-trained GNNs.
FairDrop~\cite{spinelli2021fairdrop} drops edges that cause biased predictions.
FGD~\cite{feng2023fair} generates fair distilled graphs and 
Graphair~\cite{ling2023learning} augments input graphs to mitigate sensitive information while preserving other informative features.
EDITs~\cite{Edits} iteratively adjusts node representations through fairness-aware diffusion to ensure that connected nodes across sensitive groups attain more equitable representations.
FairWalk~\cite{FairWalk} and CrossWalk~\cite{CrossWalk} aim to obtain fair representation from graph structure by modifying the random-walk sampling process.
\cite{wang2025fairness}
Some methods assume that demographic information is limited~\cite{wang2025fairness,luo2025fairness}.
DAB-GNN~\cite{lee2025disentangling} employs
a disentanglement and amplification module that isolates
and amplifies each type of bias.
FiarGT~\cite{luo2024fairgt} and FairGP~\cite{luo2025fairgp} are transformer-based methods for fair prediction on graphs.

These methods have been evaluated in terms of prediction accuracy and fairness on a limited number of datasets. 
Therefore, it is essential to identify the characteristics of these methods across diverse graph datasets.

\noindent
{\bf Graph datasets for fairness}.
Although many attributed graph datasets are publicly available, widely used graph datasets typically do not include sensitive attributes and/or target labels. A survey by Chen et al.~\cite{chen2024fairness} summarizes graphs used for fairness evaluations, such as recommendation graphs, social networks, collaboration networks, web graphs, citation networks, and synthetic (i.e., similarity) graphs. 

Among these, the social networks, Pokec-n and Pokec-z, are frequently used in group fairness studies.
Synthetic graphs, such as Credit and Bail, are essentially tabular data, and edges are synthetically generated between records in tables according to their similarities.
Other graphs are seldom used due to their small sizes and lack of node features. 
Therefore, the variety of graphs is insufficient to thoroughly evaluate fairness-aware GNN methods.
Consequently, despite the growing interest in fairness-aware GNN methods, there are few graphs available for evaluating group fairness. 

Another line of work aims to synthetically generate graphs to complement the lacking datasets. 
For example, GenCAT~\cite{maekawa2023gencat,maekawa2022beyond} and GraphWorld~\cite{palowitch2022graphworld} aim to control specific graph properties while maintaining other properties. TagGen~\cite{zhou2020data} and NetGan~\cite{bojchevski2018netgan} use neural models to imitate graph datasets and preserve the properties of the original graphs.

Recently, some methods have been proposed to generate graphs with sensitive attributes. FairGen~\cite{zheng2024fairgen} imitates the properties of given graphs concerning protected attributes. FairWire~\cite{kose2024fairwire} and FDGEN~\cite{wang2025fdgen} generate graphs that are similar but mitigate biases from the original graph.
Since these methods aim to mimic real-world graphs, newly available real-world datasets greatly enhance their usefulness.

\noindent
{\bf Fairness on knowledge graphs}.
Biases in knowledge graphs have been actively studied~\cite {kraft2022lifecycle,fisher2019measuring,pan2023large}.
They mainly focus on the biases in embeddings generated from knowledge graphs, and how to debias them~\cite{keidar2021towards} or how to find biases in knowledge graphs~\cite{sasaki2024mining}.
Since these studies are not intended for building fairness-aware models, they have not used graphs with target labels, which are necessary for supervised machine learning.

\noindent
{\bf Benchmarking fair machine learning}.
There are several benchmarking studies to evaluate the fairness of machine learning         \cite{friedler2019comparative,cardoso2019framework,islam2021can,reddy2021benchmarking,zhang2022improving,zong2022medfair,cruz2023unprocessing,han2024ffb,defrance2024abcfair,delaney2024oxonfair,jung2024counterfactually,dong2023fairness}.
Each benchmarking study and tool has different purposes, such that data types are image or text, and fair machine learning methods are inprocessing methods and/or preprocessing methods.

Among existing studies, PygDebias~\cite{dong2025fairness} and FairGAD~\cite{neo2024towards} focus on fairness in graph data.
PygDebias, which aims to evaluate both group and individual fairness, is the most closely related study to ours. It provides implementations of fairness-aware methods available up to 2022.
FairGAD, on the other hand, benchmarks fairness in outlier detection tasks, which is orthogonal to our research objective.
However, neither of these works evaluates the performance of fairness-aware models on knowledge graphs.

\section{New datasets for fairness-aware GNNs}
\label{sec:dataset}

We present new graphs generated from knowledge graphs.

\subsection{Issues in existing datasets}

Our study focuses on group fairness in node classification tasks. 
Datasets for evaluating fairness-aware GNNs should have features including labels and sensitive attributes.
However, most of the datasets rarely have both labels and sensitive attributes or even features~\cite{chen2024fairness}.

The currently often-used datasets can be divided into two types: social and synthetic graphs.
In social graphs, Pokec-n and -z are often used for fairness-aware GNNs.
These datasets only focus on regions as sensitive attributes. 
Synthetic graphs are essentially non-graph datasets because edges are synthetically generated based on the similarity of records (i.e., user profiles). 
These graphs are small-scale and less diversified in sensitive attributes and target labels.

It is essential to use various and large-scale graphs that reflect real-world issues to evaluate fairness-aware GNNs.
Knowledge graphs can be considered to reflect the real world, so these graphs are beneficial to evaluate fairness-aware GNN methods.

\subsection{Graph generation}
We describe the procedure for generating new datasets from knowledge graphs.
We have three steps: (1) extract all entities of humans from a knowledge graph, (2) extract their profiles and their relationships between humans, and (3) generate attribute graphs from extracted entities and relationships.
Our way efficiently controls what profiles and relationships are extracted before generating graphs.

We generate three new datasets from YAGO~\cite{pellissier2020yago}\footnote{YAGO4, English Wikipedia Mar 2020. \url{https://yago-knowledge.org/data/yago4/en/}}, DBpedia~\cite{auer2007dbpedia}\footnote{Dec 2022. \url{https://downloads.dbpedia.org/repo/dbpedia/mappings/mappingbased-objects/}}, and Wikidata~\cite{vrandevcic2014wikidata}\footnote{August 2024. \url{https://dumps.wikimedia.org/wikidatawiki/entities/}}, which are commonly used in the research fields.
In the generated graphs, each node represents a person, and its node features represent their profiles as a bag of attributes, where an attribute is marked as one if the person is associated with it and zero otherwise.
Edges represent relationships between nodes, such as parent–child or sibling relations.
Edges are undirected and unlabeled, as most fair GNN methods are optimized for these types of edges. 

We specify the following sensitive attributes and labels:
\begin{itemize}
\item \textbf{YAGO}.
We label nodes as either Politics/Government/Education/Academia (label 1) or Arts/Entertainment (label 0), and define sensitive attributes as nationalities, categorized into Western countries (group 1) and non-Western countries (group 0).

\item \textbf{DBpedia}.
We label nodes as either Politics/Government/Education/Academia (label 1) or Arts/Entertainment (label 0), and define sensitive attributes as nationalities, categorized into Western countries (group 1) or non-Western countries (group 0).

\item \textbf{Wikidata}. We label nodes as either Politician (label 1) or Non-politician (label 0), and define the sensitive attribute as gender, categorized into Male (group 1) and Female (group 0).

\end{itemize}

Note that YAGO and DBpedia do not contain gender information; therefore, we use nationalities as sensitive attributes.
These datasets reflect real-world contexts; thus, it is worthwhile to evaluate fairness-aware GNNs on these datasets.

\subsection{Comparison in statistics between knowledge and existing graphs}

Table~\ref{tab:dataset} shows the summary of statistics in new graphs and existing graphs that we used in our benchmarking studies.

Compared to existing graph datasets, knowledge graphs exhibit distinct statistical characteristics. Specifically, YAGO demonstrates a high degree of homophily for both sensitive attributes and labels, while the overall correlation between them remains relatively low. DBpedia, in contrast, shows a moderate level of sensitive-attribute homophily and a high level of label homophily, accompanied by a stronger correlation than that observed in YAGO. Wikidata displays the highest correlation among the three, with a strong homophily in sensitive attributes and a moderate one in labels.

These statistical differences could induce varying performance behaviors across fairness-aware GNN methods.

\begin{table*}[]
    \centering
    \caption{Dataset statistics: The first and second values within parentheses represent the sensitive attributes and target labels, respectively. Since some nodes lack either sensitive attributes and/or target labels, the sum of the four combinations---(0,0), (0,1), (1,0), and (1,1)---does not necessarily equal the total number of nodes.
``Hom. Sens'' and ``Hom. Label'' denote the homophily rates of sensitive attributes and labels, respectively; higher values indicate that connected nodes are more likely to share the same sensitive attributes or labels.
``Corr'' represents the Poisson correlation between sensitive attributes and labels.}
\vspace{-3mm}
    \label{tab:dataset}
    \resizebox{0.9\linewidth}{!}{
    \begin{tabular}{l|l|r|r|r|r|r|r|r|r|r|r}\toprule
       Type &Name  & \#Nodes & \#Feature & \#Edges & (0,0) & (0,1) & (1,0) & (1,1) &Hom. Sens & Hom. Label & Corr \\\midrule
\multirow{2}{*}{Synthetic}&Bail&18,876&18&642,616&5,457&3,860&6,315&3,244 &0.63&0.90&0.04\\
&Credit&30,000&13&304,754&5,906&21,409&730&1,955&0.81&0.52&0.08\\\midrule
\multirow{4}{*}{Social}&Pokec-n&6,185&266&36,827&2,288&1,752&1,144&1,001&0.59&0.95&0.03\\
&Pokec-z&7,659&277&48,759&2,499&2,352&1,357&1,451&0.58&0.96&0.03\\
&Pokec-n-Large&66,569&266&1,100,663&3,125&1,268&1,385&521&0.50&0.95&0.06\\
&Pokec-z-Large&67,796&277&1,303,712&3,169&1,163&1,595&801&0.46&0.95&0.00\\\midrule
\multirow{3}{*}{Knowledge}&YAGO&127,752&114&349,942&6,473&5,633&23,186&21,908&0.87&0.97&0.02\\
&DBpedia&152,035&478&426,851&707&850&559&1,141&0.64&0.97&0.13\\
&Wikidata&1,108,347&415&4,857,725&12,342&3,342&16,119&53,922&0.80&0.44&0.46\\
\bottomrule
    \end{tabular}
    }
\end{table*}

\section{Benchmark}
\label{sec:experiment}
We present the benchmarks on node classification tasks in terms of prediction accuracy and fairness.
We designed the benchmark for the following questions:
{\bf Q1}. How is the difference in overall trends across the dataset types?
{\bf Q2}.  How does the performance of fairness-aware methods vary across the datasets?
{\bf Q3}. What are the important factors among fairness-aware methods, backbone models, and early stopping conditions?
{\bf Q4}. How are fairness-aware methods sensitive to backbone models and early stopping conditions?
{\bf Q5}. How scalable are fairness-aware GNN methods?


We implemented all methods using Python3 and used a server with NVIDIA  Quadro RTX 8000 GPU and 48 GB GPU memory to run all methods.

\subsection{Setting}

We here describe datasets, baselines, metrics, and hyperparameters.
Since we provide graphs, data splitting, implementations of baselines and metrics, and hyperparameters, the results are reproducible and reused for future studies.

\noindent 
{\bf Datasets}.
We use the six existing common graphs and three new knowledge graphs in Table~\ref{tab:dataset}.
We divide a set of nodes into train/validation/test in 0.6/0.2/0.2 following existing works~\cite{chien2020adaptive,pei2020geom}.

\noindent 
{\bf Baselines}.
We use four fair inprocessing methods;
\textbf{FairGNN}~\cite{FairGNN}, \textbf{NIFTY} ~\cite{Nifty},  \textbf{FairSIN} \cite{yang2024fairsin}, and \textbf{FairGB}~\cite{li2024rethinking} and three preprocessing methods; \textbf{Undersampling}, \textbf{FairDrop}~\cite{spinelli2021fairdrop}, and \textbf{BIND}~\cite{BIND}, with a \textbf{Vanilla} method.
We employ four common GNN models as their backbones; GCN~\cite{GCN},  GAT~\cite{GAT}, SAGE \cite{GraphSage}, and H2GCN~\cite{zhu2020beyond}\footnote{The latest methods are FairSIN and FairGB published in 2024. 
We note that it is hard to thoroughly reimplement and evaluate existing methods (in particular, those published in 2025) because our benchmarking studies incur significant computational costs.
We selected standard and the SOTA methods; for example, since FairSIN outperforms FairVGNN~\cite{FairVGNN} and EDITS~\cite{Edits}, we skipped them.}.

FairGB is incompatible with H2GCN because H2GCN requires a constant number of nodes during training, whereas FairGB dynamically changes the node set.
BIND did not finish within 24 hours except for Pokec-n and Pokec-z. 
NIFTY and FairGB caused out-of-memory errors in some cases.

\noindent 
{\bf Metrics}.
We evaluate the performance of baselines in terms of prediction accuracy and fairness metrics. We report their averages and variances over five runs in different data splits.

We use three prediction accuracy metrics: Accuracy (ACC), AUC, and F1.
According to prior studies~\cite{beutel2017data, TheVariationalFairAutoencoder}, we use $\Delta SP$~\cite{dwork2012fairness} and $\Delta Eop$~\cite{hardt2016equality} to quantitatively evaluate statistical parity and equal opportunity.
$\Delta SP = |P(\hat{y} = 1\mid s = 0) - P(\hat{y} = 1\mid s = 1)|$ and $\Delta Eop = |P(\hat{y} = 1 \mid y = 1, s = 0) - P(\hat{y} = 1 \mid y = 1, s = 1)|$, where $y \in \{0, 1\}$ represents the label, variable $s \in \{0, 1\}$ represents the sensitive attribute, and $\hat{y} \in \{0, 1\}$ indicates prediction output.
Both $\Delta SP$ and $\Delta Eop$ are ``smaller-is-better'' metrics ($\downarrow$), and ACC, AUC, and F1 are ``larger-is-better'' metrics ($\uparrow$).


\noindent
{\bf Early stopping condition}.
To evaluate the trade-offs between prediction accuracy and fairness, we use three types of early stopping conditions to maximize (i) accuracy, (ii) F1, and (iii) (ACC + F1) - 0.5($\Delta SP$  + $\Delta Eop$ ), namely Acc, F1, and Mixed, respectively.

\noindent
{\bf Hyper-parameters}.
We perform hyperparameter tuning for each run. 
The hyperparameter search space is described in  Appendix~\ref{sec:hyperparameters}.

\subsection{Experimental results}



\begin{figure*}[t!]
    \centering
    \begin{minipage}{\textwidth}
        \centering
        \includegraphics[width=1.0\linewidth]{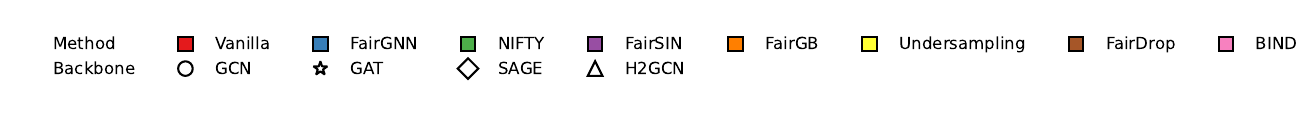}
        \vspace{-8mm}
    \end{minipage}
    \begin{minipage}{0.3\textwidth}
        \includegraphics[width=1.0\linewidth]{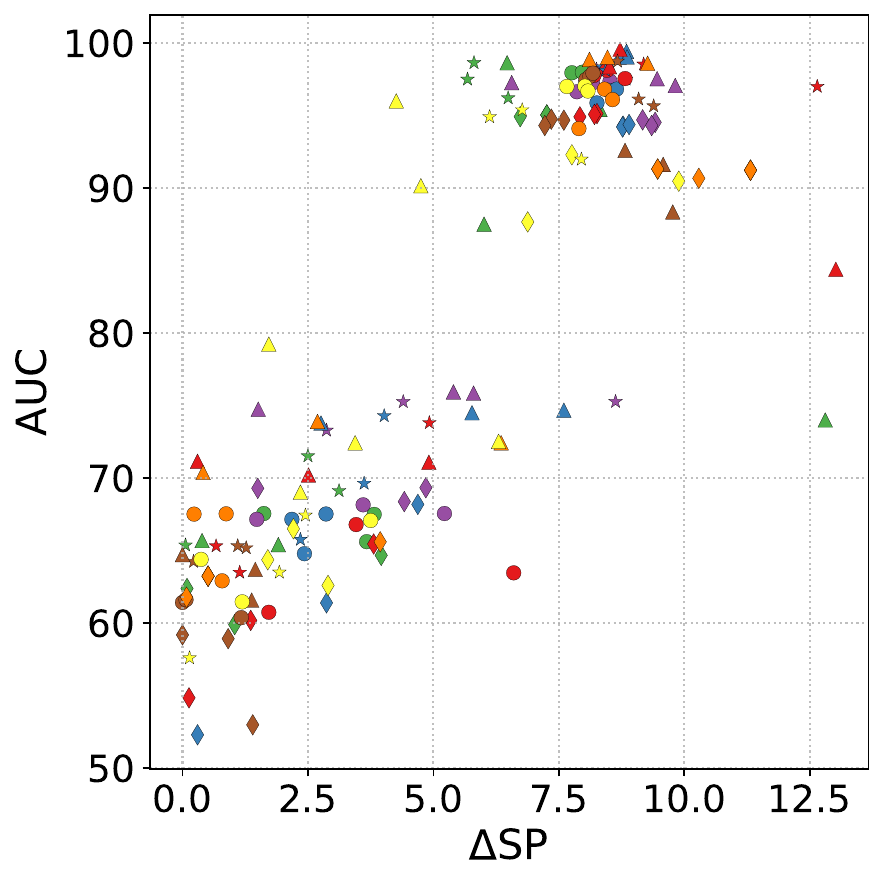}
        \vspace{-7mm}
        \subcaption{$\Delta$SP ($\downarrow$) and AUC ($\uparrow$) in synthetic graphs}
    \end{minipage}
        \begin{minipage}{0.3\textwidth}
        \includegraphics[width=1.0\linewidth]{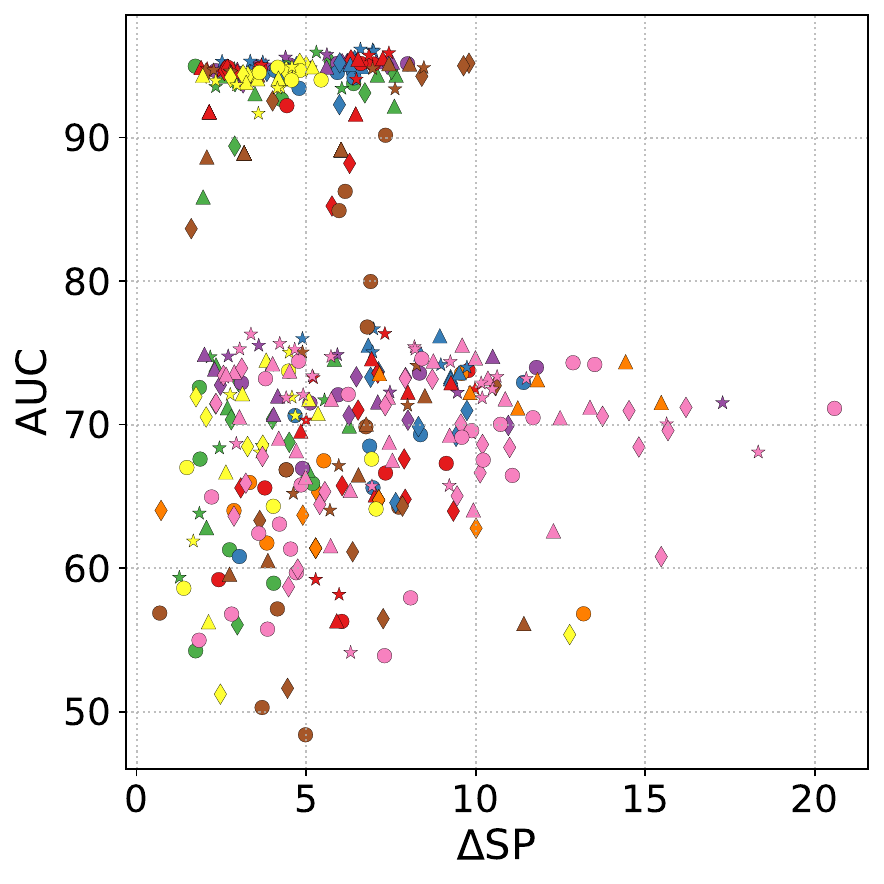}
        \vspace{-7mm}
        \subcaption{$\Delta$SP ($\downarrow$) and AUC ($\uparrow$) in social graphs}
    \end{minipage}
        \begin{minipage}{0.3\textwidth}
        \includegraphics[width=1.0\linewidth]{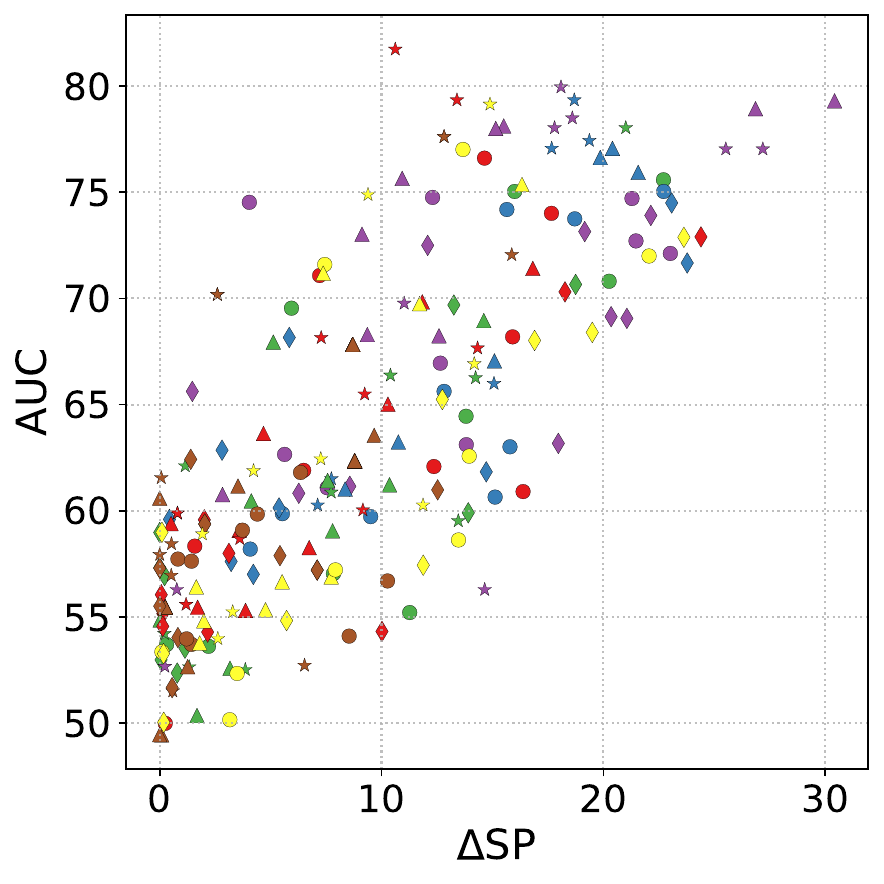}
        \vspace{-7mm}
        \subcaption{$\Delta$SP ($\downarrow$) and AUC ($\uparrow$) in knowledge graphs}
    \end{minipage}

   \begin{minipage}{0.3\textwidth}
        \includegraphics[width=1.0\linewidth]{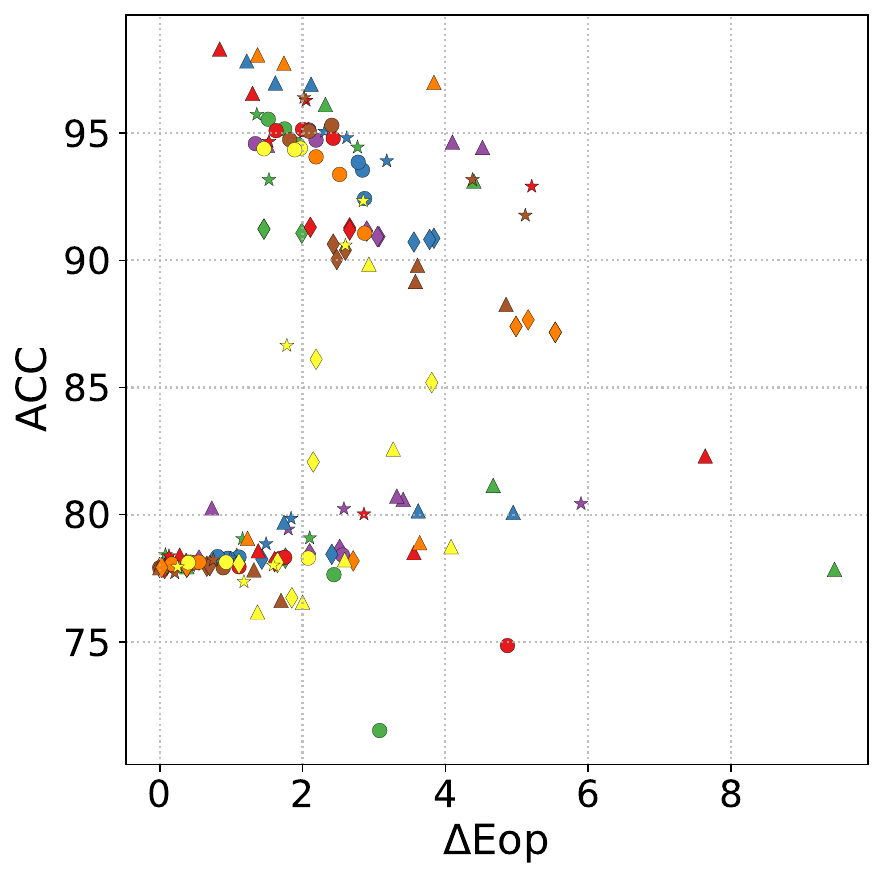}
        \vspace{-7mm}
        \subcaption{$\Delta$Eop ($\downarrow$) and ACC ($\uparrow$) in synthetic graphs}
    \end{minipage}
        \begin{minipage}{0.3\textwidth}
        \includegraphics[width=1.0\linewidth]{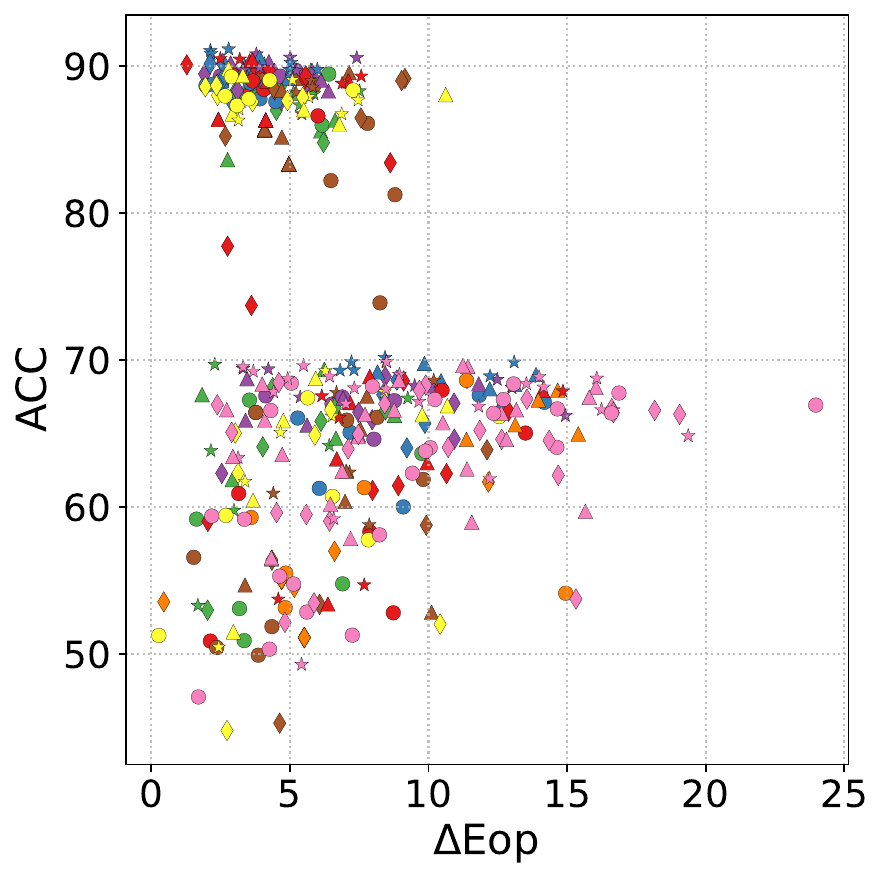}
        \vspace{-7mm}
        \subcaption{$\Delta$Eop ($\downarrow$) and ACC ($\uparrow$) in social graphs}
    \end{minipage}
        \begin{minipage}{0.3\textwidth}
        \includegraphics[width=1.0\linewidth]{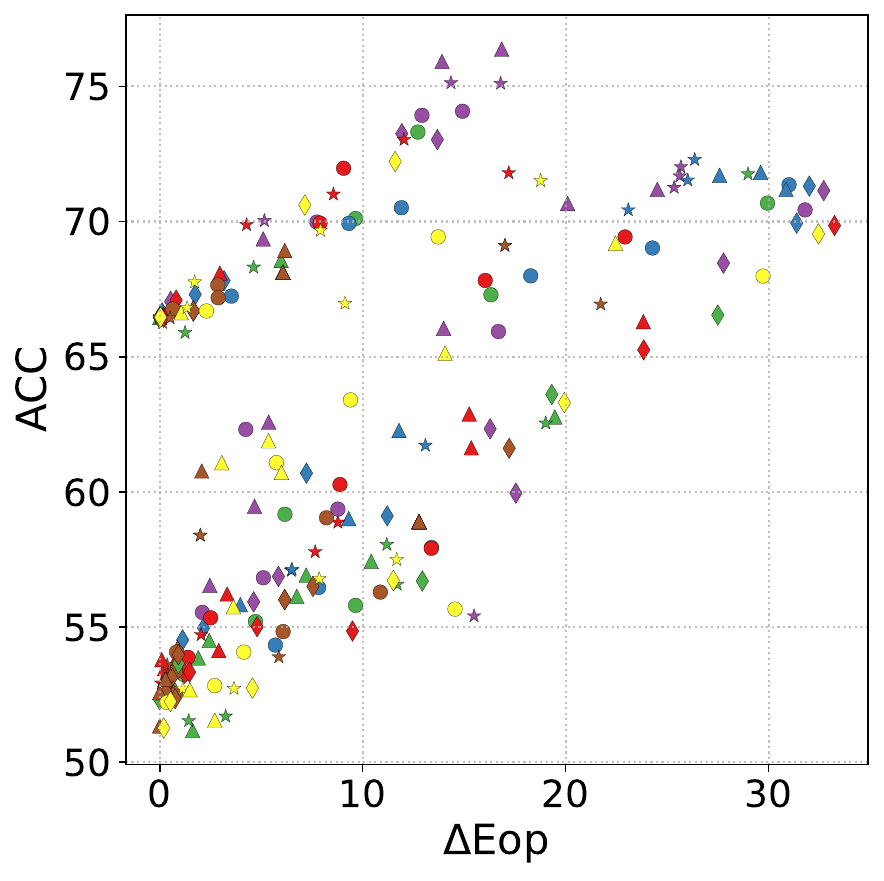}
        \vspace{-7mm}
        \subcaption{$\Delta$Eop ($\downarrow$) and ACC ($\uparrow$) in knowledge graphs}
    \end{minipage}
    
    \vspace{-3mm}
    \caption{The overall trends of trade-offs in dataset types.}
    \label{fig:result_tradeoff_main}
    \vspace{-4mm}
\end{figure*}

\noindent
{\bf Q1. Overall trend difference between knowledge and existing graphs}.
Figure~\ref{fig:result_tradeoff_main} shows the overall trade-offs between prediction accuracy and fairness metrics in three dataset types. Each figure represents different metrics of prediction accuracy and fairness.
Each plot in the figures indicates the performance of a triplet of method, model, and early stopping condition.
Since each figure includes multiple graphs (e.g., Credit and Bail in the synthetic), distinct clusters emerge reflecting varying dataset difficulty.
The trade-offs between other metrics are in Appendix~\ref{sec:detailedresults}.

From Figure~\ref{fig:result_tradeoff_main}, we can see that knowledge graphs exhibit significant trade-offs between prediction accuracy and fairness metrics compared to social and synthetic graphs.
This indicates that, in social and synthetic graphs, fairness metrics can be improved while preserving prediction accuracy if we carefully select methods, models, and early stopping conditions.
On the other hand, knowledge graphs are challenging to improve fairness metrics while preserving prediction accuracy.

Table~\ref{tab:tradeoff_pearson_main} shows the Pearson correlation and p-values in each dataset.
Knowledge graphs statistically exhibit stronger correlations between prediction accuracy and fairness metrics, indicating that improvements in prediction accuracy are often accompanied by deteriorations in fairness.
On the other hand, Bail, Pokec-n-Large, and Pokec-z-Large have negative values, indicating that improving prediction accuracy often contributes to the improvement in fairness.  
In addition, we can see that different pairs of metrics have different degrees of correlation.

These results yield three key observations.
First, different prediction accuracy and fairness have different trade-offs (see in Appendix~\ref{sec:detailedresults} in additional results).
Second, knowledge graphs may be more challenging to achieve both prediction accuracy and fairness metrics compared to existing graphs.
Third, existing methods may not be suitable for knowledge graphs because they validate their performance by using knowledge graphs. 
We may need to find ways to balance prediction accuracy and fairness, depending on our specific applications.




\noindent
{\bf Q2. Performance difference of fairness-aware GNNs}.
We analyze the suitability of methods for different dataset types, considering Pareto optimality and average ranking.

Table~\ref{table:result_GCN_ACC} shows the performance of fairness-aware methods in a standard setting, which is a GCN backbone model with ACC early stopping condition~\cite {dong2025fairness}.
Compared to methods, vanilla, NIFTY, and FairSIN often achieve better prediction accuracy, and Undersampling and FairDrop achieve better fairness metrics.
Fairness-aware methods often exhibit higher prediction performance but lower fairness compared to Vanilla, which was not their original intention.

\begin{table}[h!]
\caption{Pearson correlation between prediction accuracy and fairness metrics. Larger positive values indicate a clearer trade-off between prediction accuracy and fairness. Values in parentheses denote $p$-values.}
\label{tab:tradeoff_pearson_main}
\vspace{-3mm}
\begin{tabular}{lcc}
\toprule
Dataset & AUC--$\Delta$SP & ACC--$\Delta$Eop \\
\midrule
Credit & 0.62 (0.000) & 0.13 (0.235) \\
Bail & 0.05 (0.643) & -0.52 (0.000) \\
Pokec-n & 0.38 (0.000) & 0.62 (0.000) \\
Pokec-z & 0.22 (0.034) & 0.45 (0.000) \\
Pokec-n-Large & -0.23 (0.032) & -0.04 (0.707) \\
Pokec-z-Large & 0.05 (0.668) & -0.22 (0.046) \\
DBpedia & 0.70 (0.000) & 0.75 (0.000) \\
YAGO & 0.68 (0.000) & 0.84 (0.000) \\
Wikidata & 0.87 (0.000) & 0.65 (0.000) \\
\bottomrule
\end{tabular}
\vspace{-5mm}
\end{table}

\begin{figure}[ttt]
    \centering
    \begin{minipage}{.95\linewidth}
        \includegraphics[width=1.0\linewidth]{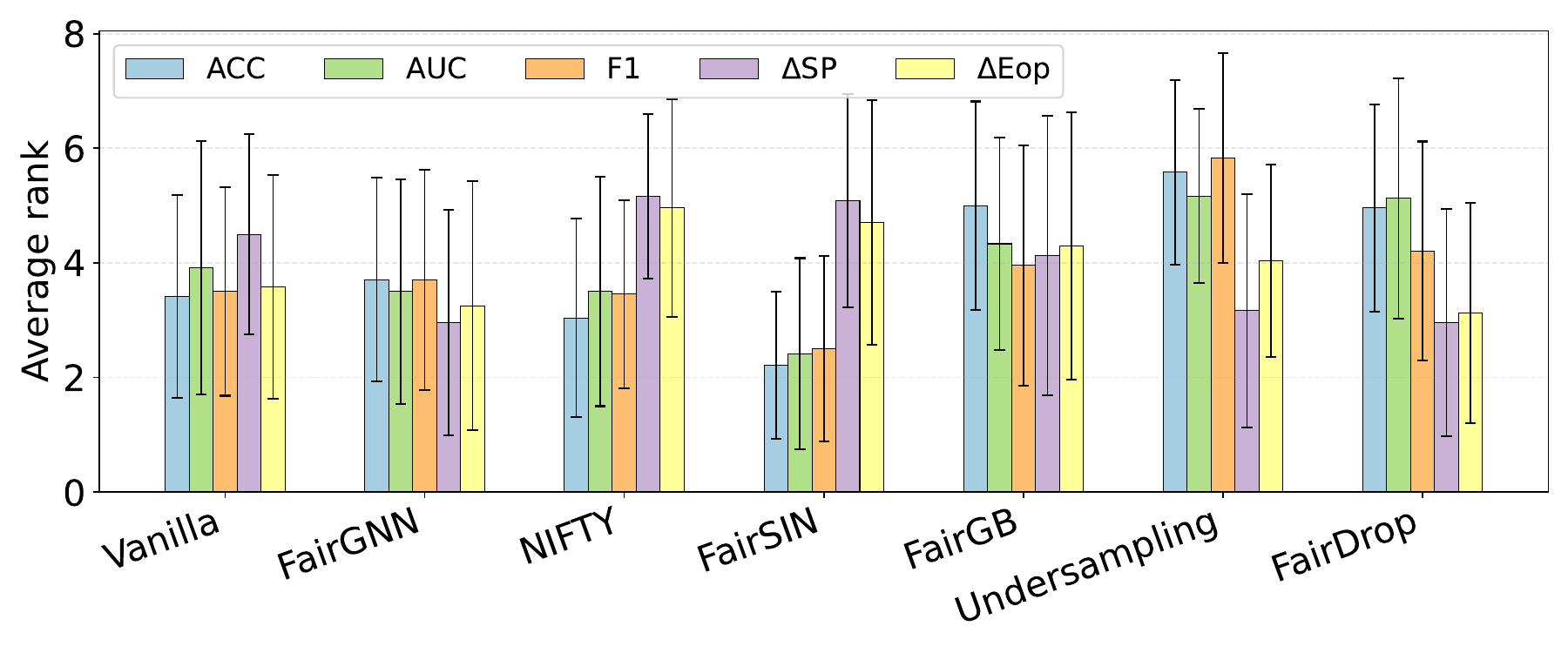}
        \vspace{-7mm}

        \subcaption{Synthetic graphs}
    \end{minipage}
        \begin{minipage}{0.95\linewidth}        \includegraphics[width=1.0\linewidth]{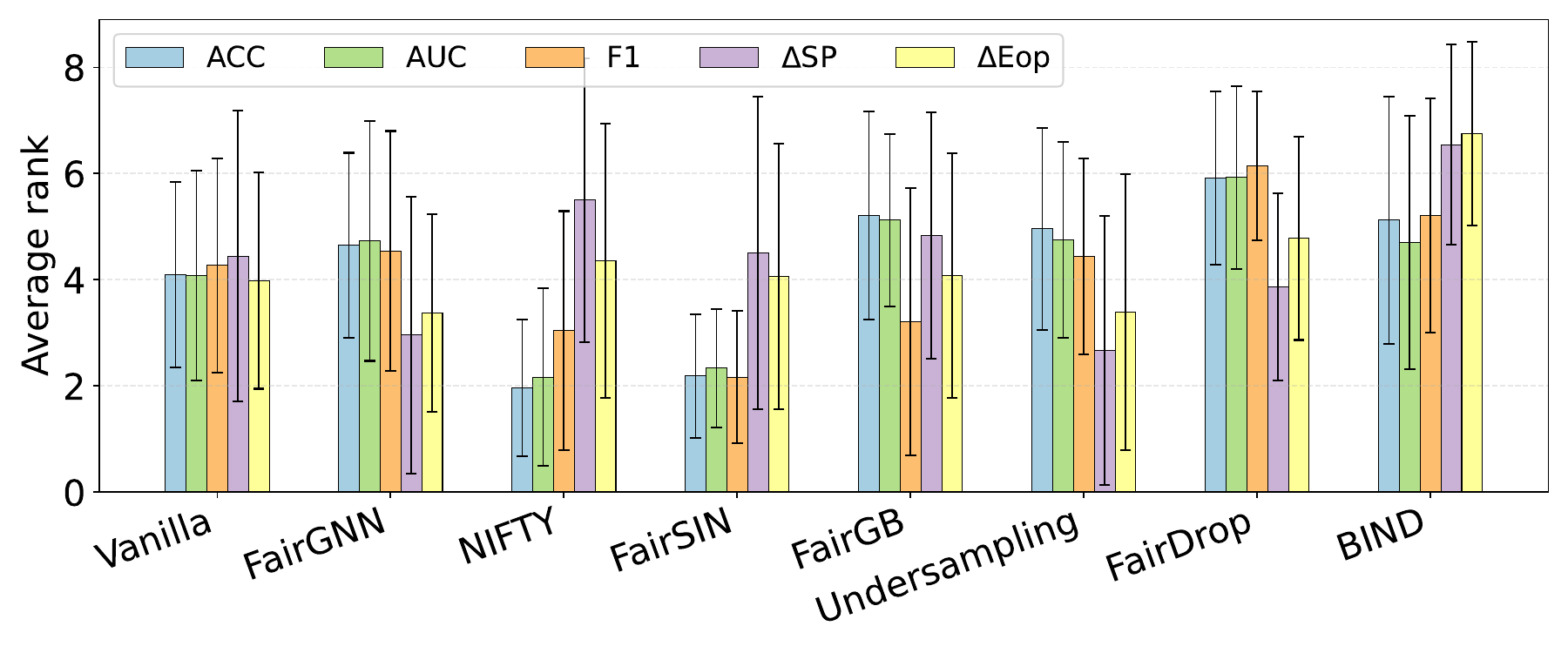}
        \vspace{-7mm}

        \subcaption{Social graphs}
    \end{minipage}
        \begin{minipage}{0.95\linewidth}        \includegraphics[width=1.0\linewidth]{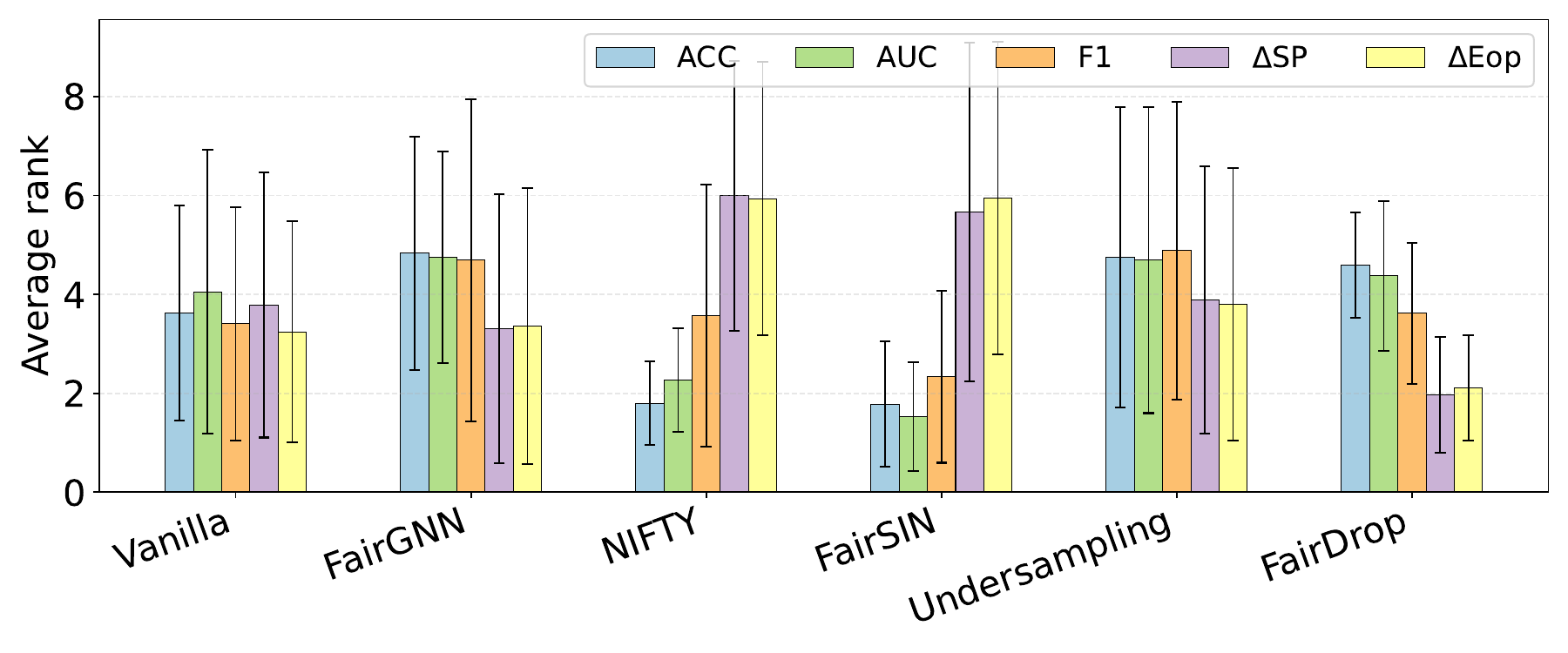}
        \vspace{-7mm}
        \subcaption{Knowledge graphs}
    \end{minipage}
    \vspace{-3mm}
    \caption{Average ranking in different dataset types}
    \label{fig:result_ranking}
    \vspace{-7mm}
\end{figure}

Figure~\ref{fig:result_ranking} shows the average ranking of fairness-aware methods across dataset types.
We can see that NIFTY and FairSIN achieve higher rankings in prediction accuracy, while FairGNN, Undersampling, and FairDrop have better rankings in fairness metrics, exhibiting similar tendencies as shown in Table~\ref{table:result_GCN_ACC}.

\begin{table*}[!h]
\caption{Prediction accuracy and fairness on GCN backbone and ACC early stopping. Bold and underlined values indicate the best and second-best methods, respectively. OOM and DNF indicate ``out-of-memory errors'' and ``did not finish within 24 hours'', respectively.}\label{table:result_GCN_ACC}
\vspace{-3mm}
\begin{center}
\resizebox{1.0\linewidth}{!}{
\begin{tabular}{ c | c | c c c c c c c c c } \hline
 Metric  & Method & Credit & Bail & Pokec-n & Pokec-z & Pokec-n-Large & Pokec-z-Large & YAGO & DBpedia & Wikidata \\ \hline\hline
\multirow{8}{*}{ACC}&Vanilla & 78.15±0.00 & \textbf{91.28±0.00} & 66.53±0.02 & 68.58±0.01 & \textbf{89.13±0.01} & \underline{90.07±0.01} & 69.85±0.01 & 54.85±0.12 & 67.09±0.01 \\
&FairGNN & 78.29±0.00 & \underline{91.23±0.00} & 66.61±0.01 & 68.12±0.01 & 84.78±0.48 & 87.64±0.20 & 66.54±0.25 & 56.70±0.05 & 66.50±0.00 \\
&NIFTY & \underline{78.44±0.00} & 90.87±0.00 & \underline{66.79±0.01} & 68.64±0.00 & \underline{88.98±0.00} & \textbf{90.13±0.00} & \textbf{71.31±0.00} & \underline{59.11±0.04} & 67.81±0.03 \\
&FairSIN & \textbf{78.63±0.00} & 91.15±0.00 & \textbf{67.07±0.02} & \textbf{68.88±0.01} & 87.67±0.01 & 89.48±0.01 & \underline{71.14±0.00} & \textbf{59.95±0.01} & \textbf{73.03±0.01} \\
&FairGB & 78.18±0.00 & 87.66±0.00 & 61.71±0.17 & 57.01±0.29 & OOM & OOM & OOM & OOM & OOM \\
&Undersampling & 78.12±0.00 & 85.19±0.07 & 66.63±0.04 & 65.04±0.45 & 87.86±0.01 & 87.99±0.00 & 69.54±0.01 & 56.72±0.14 & \underline{72.22±0.00} \\
&FairDrop & 77.99±0.00 & 90.40±0.00 & 63.90±0.32 & \underline{68.75±0.01} & 85.22±0.44 & 89.14±0.01 & 61.61±0.62 & 56.01±0.18 & 66.69±0.00 \\
&BIND & DNF & DNF & 66.31±0.06 & 64.06±0.41 & DNF & DNF & DNF & DNF & DNF \\
\hline
\multirow{8}{*}{F1}&Vanilla & 65.44±0.27 & \textbf{95.12±0.00} & 67.61±0.05 & \underline{73.59±0.00} & \textbf{94.66±0.00} & \textbf{95.41±0.00} & 72.90±0.02 & 54.31±0.41 & 59.57±0.23 \\
&FairGNN & 64.66±0.48 & \underline{95.03±0.00} & 68.77±0.04 & 73.01±0.01 & 89.40±0.98 & 93.11±0.19 & 70.66±0.25 & 59.90±0.14 & 56.94±0.01 \\
&NIFTY & \underline{68.17±0.02} & 94.26±0.00 & \underline{69.18±0.02} & \textbf{73.68±0.00} & \underline{94.61±0.00} & \underline{95.36±0.00} & \textbf{74.50±0.02} & \underline{61.83±0.12} & 60.13±0.24 \\
&FairSIN & \textbf{69.32±0.01} & 94.30±0.00 & \textbf{70.31±0.01} & 73.33±0.01 & 93.81±0.00 & 95.15±0.00 & \underline{73.91±0.00} & \textbf{63.17±0.04} & \textbf{69.06±0.01} \\
&FairGB & 65.59±0.10 & 90.66±0.00 & 62.78±0.65 & 65.28±0.18 & OOM & OOM & OOM & OOM & OOM \\
&Undersampling & 66.49±0.04 & 90.46±0.03 & 68.57±0.06 & 71.94±0.06 & 94.19±0.00 & 94.62±0.00 & 72.88±0.03 & 57.43±0.53 & \underline{68.40±0.00} \\
&FairDrop & 52.97±1.46 & 94.67±0.01 & 64.32±0.85 & 72.72±0.01 & 83.65±4.77 & 95.18±0.00 & 60.98±1.63 & 57.20±0.43 & 59.37±0.03 \\
&BIND & DNF & DNF & 68.44±0.07 & 68.41±0.57 & DNF & DNF & DNF & DNF & DNF \\
\hline
\multirow{8}{*}{AUC}&Vanilla & 87.50±0.00 & \underline{87.48±0.01} & 53.94±0.17 & \underline{64.55±0.03} & \textbf{80.96±0.01} & \textbf{82.30±0.02} & 70.43±0.00 & \underline{65.60±0.14} & 80.07±0.00 \\
&FairGNN & 87.57±0.00 & 87.46±0.01 & 54.43±0.07 & 63.79±0.02 & 72.02±2.30 & 77.72±0.64 & 68.51±0.07 & 60.61±0.37 & 79.85±0.00 \\
&NIFTY & 87.52±0.00 & 86.92±0.00 & \textbf{56.93±0.03} & \textbf{65.05±0.02} & \underline{80.48±0.01} & \underline{82.16±0.01} & \textbf{71.82±0.00} & 62.83±0.05 & 80.13±0.00 \\
&FairSIN & \textbf{87.66±0.00} & \textbf{87.60±0.01} & \underline{56.44±0.17} & 63.36±0.01 & 77.87±0.01 & 81.20±0.05 & \underline{71.67±0.00} & 63.14±0.03 & \textbf{82.31±0.00} \\
&FairGB & 87.44±0.00 & 82.70±0.00 & 42.07±4.50 & 49.83±6.27 & OOM & OOM & OOM & OOM & OOM \\
&Undersampling & 87.54±0.00 & 80.62±0.06 & 52.31±0.47 & 51.80±6.71 & 80.02±0.02 & 80.33±0.01 & 70.19±0.04 & 65.31±0.13 & \underline{81.78±0.00} \\
&FairDrop & \underline{87.59±0.00} & 86.03±0.01 & 44.27±2.87 & 64.23±0.04 & 63.07±9.95 & 79.43±0.04 & 70.00±0.03 & \textbf{67.24±0.04} & 79.71±0.00 \\
&BIND & DNF & DNF & 54.23±0.06 & 50.21±6.32 & DNF & DNF & DNF & DNF & DNF \\
\hline
\multirow{8}{*}{$\Delta$SP}&Vanilla & 3.81±0.08 & 8.26±0.05 & 7.89±0.04 & 7.11±0.02 & \underline{2.59±0.03} & 6.32±0.03 & 24.40±0.01 & \underline{10.02±1.51} & \underline{2.01±0.07} \\
&FairGNN & 3.96±0.06 & \textbf{7.26±0.04} & \underline{4.50±0.05} & \underline{3.00±0.00} & 2.89±0.02 & 6.73±0.02 & \underline{18.75±0.51} & 13.91±0.48 & \textbf{0.22±0.00} \\
&NIFTY & 4.69±0.02 & 8.78±0.06 & 9.42±0.14 & 7.13±0.01 & 3.84±0.02 & 6.43±0.04 & 23.08±0.01 & 14.72±0.19 & 5.38±0.38 \\
&FairSIN & 4.85±0.13 & 9.35±0.05 & 8.00±0.29 & 6.48±0.08 & 3.14±0.05 & \underline{6.21±0.03} & 22.14±0.00 & 17.97±0.08 & 21.06±0.16 \\
&FairGB & 3.94±0.05 & 10.29±0.02 & 10.01±0.33 & 5.33±0.28 & OOM & OOM & OOM & OOM & OOM \\
&Undersampling & \underline{2.21±0.04} & 9.89±0.06 & \textbf{3.72±0.05} & \textbf{1.75±0.01} & 2.77±0.03 & \textbf{4.41±0.03} & 23.63±0.15 & 11.88±0.94 & 19.50±0.03 \\
&FairDrop & \textbf{1.40±0.05} & \underline{7.60±0.02} & 7.84±0.20 & 10.58±0.09 & \textbf{1.61±0.01} & 9.80±0.05 & \textbf{12.53±1.10} & \textbf{7.10±0.67} & 2.03±0.12 \\
&BIND & DNF & DNF & 14.81±0.07 & 11.00±0.55 & DNF & DNF & DNF & DNF & DNF \\
\hline
\multirow{8}{*}{$\Delta$Eop}&Vanilla & \underline{1.61±0.02} & 2.66±0.02 & 12.89±0.14 & 9.10±0.06 & 5.54±0.09 & \textbf{1.28±0.01} & 33.25±0.01 & \underline{9.50±1.39} & \underline{0.81±0.01} \\
&FairGNN & 1.76±0.01 & \textbf{1.46±0.02} & \underline{8.45±0.27} & \underline{4.38±0.02} & 6.21±0.10 & 3.03±0.05 & \underline{27.50±0.60} & 12.94±0.66 & \textbf{0.19±0.00} \\
&NIFTY & 2.41±0.01 & 3.84±0.01 & 12.66±0.42 & 8.86±0.08 & \underline{3.59±0.05} & \underline{2.12±0.02} & 32.01±0.01 & 11.21±0.35 & 3.17±0.17 \\
&FairSIN & 2.52±0.06 & 2.90±0.02 & 10.92±0.44 & 8.44±0.10 & 4.69±0.08 & 4.44±0.05 & 32.72±0.01 & 17.55±0.18 & 13.68±0.16 \\
&FairGB & 2.71±0.02 & 5.16±0.03 & 12.16±0.47 & 6.61±0.29 & OOM & OOM & OOM & OOM & OOM \\
&Undersampling & 1.65±0.03 & 3.81±0.04 & \textbf{6.47±0.22} & \textbf{3.03±0.05} & 5.46±0.09 & 2.39±0.04 & 32.46±0.27 & 11.51±0.90 & 11.60±0.06 \\
&FairDrop & \textbf{0.70±0.01} & \underline{2.60±0.03} & 12.11±0.56 & 11.39±0.08 & \textbf{2.67±0.04} & 9.15±0.06 & \textbf{17.22±1.97} & \textbf{6.16±0.54} & 1.66±0.08 \\
&BIND & DNF & DNF & 19.06±0.12 & 10.71±0.40 & DNF & DNF & DNF & DNF & DNF \\
\hline
\end{tabular}
}
\end{center}
\vspace{-3mm}
\end{table*}

Table~\ref{tab:pareto_counts_method} shows the number of Pareto-optimal points in each dataset among five metrics.
We can observe that different datasets have different trends.
Vanilla, FairGNN, NIFTY, FairSIN, FairGB, and Undersampling are often selected as Pareto-optimal in multiple datasets.
FairDrop is effective in knowledge graphs.
Vanilla does not achieve good performance in Pokec-n and Pokec-z, indicating that fairness-aware methods improve both prediction accuracy and fairness metrics compared to Vanilla.
NIFTY does not work well for synthetic graphs, and FairGB is effective in synthetic graphs and Pokec-n.
FairGNN achieves high performance in both prediction accuracy and fairness on the Bail dataset, obtaining the second-best results in ACC and F1, and the best results in $\Delta$SP and $\Delta$Eop.

These results indicate that suitable methods are different across datasets.
They show three observations. First, it is hard to achieve high rankings in both accuracy prediction and fairness metrics.
Second, fairness-aware methods sometimes enhance prediction accuracy while degrading fairness metrics, which contradicts their intended purpose.
Third, if we carefully select methods, models, and early stopping conditions, biased prediction can be mitigated while keeping accuracy in several graphs.

\noindent
{\bf Q3. Important factors}.
We here analyze the impact on method, model, and early stopping conditions.
Each factor has its own purpose and sometimes works in an unexpected direction, as we showed in Q2, so it is not obvious what the impact of their combination is on performance.
In particular, early stopping conditions empirically have a significant impact on both prediction accuracy and fairness.
Each paper uses different early stopping conditions. For example, the original implementation of FairSIN controls early stopping to balance prediction accuracy and fairness through its hyper-parameters.

Table~\ref{tab:shapley_by_datasettype} shows the Shapley relative importance values in \% of fairness-aware methods, backbone models, and early stopping conditions, for explaining the variance in different evaluation metrics.
Each cell indicates the proportion of the variation in the metric that can be attributed to the factor.

Across all dataset types and metrics, methods consistently account for a large share of importance.
This suggests that the choice of fairness-aware methods has the strongest influence on outcomes.
For the synthetic graphs, the model becomes almost as important as the method in terms of prediction accuracy, while early stopping conditions are crucial for fairness metrics. This indicates that the model choice plays an important role in prediction accuracy, but not fairness, while selecting early stopping conditions largely affects fairness metrics.
In social graphs, the importance of methods is dominant in all metrics. This might be due to the fact that methods and models are often evaluated using Pokec datasets.
In the knowledge graphs, the early stopping conditions contribute substantially compared with other dataset types.
This indicates that training dynamics strongly affect fairness and accuracy in knowledge graphs.

These results show that the method influences evaluation metrics most strongly, but early stopping conditions are the most important for fairness in synthetic and knowledge graphs.
Therefore, early stopping conditions constitute important hyperparameters affecting fairness in graph learning, and their impact warrants further investigation in future work.



\noindent
{\bf Q4. Sensitivity of fairness-aware method to model and early stopping condition}.
We validate how fairness-aware methods are sensitive to backbone models and early stopping conditions by evaluating the ratios of performance changes.
Tables~\ref{tab:std_model} and \ref{tab:std_earlystop} show the standard deviations of ratios of the performance changes when backbone models and early stopping conditions change, respectively.
If the standard deviation is small, the methods have a small impact even if backbone models or early stopping conditions change.

From Table~\ref{tab:std_model}, we observe that FairSIN and FairDrop exhibit relatively small standard deviations on average, whereas Vanilla, FairGNN, and FairGB show comparatively larger variations.
In social and knowledge graphs, FairDrop and Undersampling demonstrate particularly large deviations, respectively.
These findings indicate that Vanilla, FairGNN, and FairGB are more sensitive to model variations, and the degree of sensitivity differs across dataset types.

Compared with model variations, under different early stopping conditions, NIFTY and FairSIN maintain smaller deviations.
The standard deviations of fairness metrics under early stopping conditions are generally larger than those under model variations, suggesting that early stopping has a greater impact on fairness performance than model choice.

Overall, these results suggest that certain methods, such as Vanilla and FairGNN, are sensitive to both model choice and early stopping conditions. Therefore, evaluating their performance under a single experimental setting is insufficient for a fair comparison.

\begin{table}[t!]
\centering
\caption{The number of Pareto-optimal points}
\label{tab:pareto_counts_method}
\vspace{-3mm}
\setlength{\tabcolsep}{6pt}
\resizebox{1.0\linewidth}{!}{
\begin{tabular}{lrrrrrrrr}
\toprule
Dataset & Vanilla & FairGNN & NIFTY & FairSIN & FairGB & Und. & FairDrop & BIND \\
\midrule
Credit & 1 & 2 & 0 & \textbf{4} & 3 & 0 & 3 & - \\
Bail & 2 & \textbf{3} & 0 & 1 & 2 & 2 & 0 & - \\
Pokec-n & 0 & 5 & 3 & 5 & \textbf{6} & 4 & 3 & 0 \\
Pokec-z & 0 & \textbf{6} & 3 & \textbf{6} & 1 & 4 & 1 & 0 \\
Pokec-n-Large & \textbf{8} & 3 & \textbf{8} & 7 & - & 1 & 3 & - \\
Pokec-z-Large & 2 & 1 & \textbf{7} & 5 & - & 6 & 0 & - \\
DBpedia & 7 & 1 & 8 & 8 & - & 1 & \textbf{17} & - \\
YAGO & 1 & 1 & 0 & 1 & - & 2 & \textbf{6} & - \\
Wikidata & 9 & 8 & 2 & 9 & - & 3 & \textbf{13} & - \\
\bottomrule
\end{tabular}
}
\end{table}


\begin{table}[t]
\centering
\caption{Shapley relative importance (\%).}
\label{tab:shapley_by_datasettype}
\vspace{-3mm}
\setlength{\tabcolsep}{6pt}
\begin{tabular}{llrrr}
\toprule
\multirow{2}{*}{Dataset type} & \multirow{2}{*}{Metric} & \multicolumn{3}{c}{Shapley relative importance (\%)} \\
\cmidrule(lr){3-5}
 & & Method & Model & Early stop \\
\midrule
\multirow{5}{*}{Synthetic} & ACC & \textbf{53.9\%} & 43.7\% & 2.4\% \\
 & AUC & 44.4\% & \textbf{51.8\%} & 3.8\% \\
 & F1 & 47.5\% & \textbf{51.7\%} & 0.7\% \\
 & $\Delta$SP & \textbf{55.1\%} & 6.5\% & 38.4\% \\
 & $\Delta$Eop & 12.3\% & 39.2\% & \textbf{48.5\%} \\
\midrule
\multirow{5}{*}{Social} & ACC & \textbf{78.0\%} & 14.6\% & 7.4\% \\
 & AUC & \textbf{77.2\%} & 17.7\% & 5.1\% \\
 & F1 & \textbf{74.1\%} & 22.1\% & 3.9\% \\
 & $\Delta$SP & \textbf{84.4\%} & 0.9\% & 14.7\% \\
 & $\Delta$Eop & \textbf{69.1\%} & 1.1\% & 29.7\% \\
\midrule
\multirow{5}{*}{Knowledge} & ACC & \textbf{46.6\%} & 9.3\% & 44.2\% \\
 & AUC & \textbf{69.7\%} & 13.1\% & 17.2\% \\
 & F1 & \textbf{58.4\%} & 31.4\% & 10.3\% \\
 & $\Delta$SP & \textbf{48.5\%} & 4.0\% & 47.4\% \\
 & $\Delta$Eop & 48.3\% & 2.3\% & \textbf{49.5\%} \\
\bottomrule
\end{tabular}

\end{table}

\begin{table*}[t]
\centering
\caption{Standard deviation of performance change ratio in model. `---' indicates no results.}
\label{tab:std_model}
\vspace{-3mm}
\setlength{\tabcolsep}{6pt}
\resizebox{1.0\linewidth}{!}{
\begin{tabular}{lrrrrrrrrrrrrrrr}
\toprule
\multirow{2}{*}{Method} & \multicolumn{5}{c}{Synthetic} & \multicolumn{5}{c}{Social} & \multicolumn{5}{c}{Knowledge} \\
\cmidrule(lr){2-6}\cmidrule(lr){2-6}\cmidrule(lr){7-11}\cmidrule(lr){12-16}
 & ACC & AUC & F1 & $\Delta$SP & $\Delta$Eop & ACC & AUC & F1 & $\Delta$SP & $\Delta$Eop & ACC & AUC & F1 & $\Delta$SP & $\Delta$Eop \\
\midrule
Vanilla & 3.40 & 7.74 & 4.88 & 286.36 & 426.58 & 8.23 & 9.57 & 16.19 & 26.84 & 67.77 & 6.26 & 7.99 & 10.63 & 2281.49 & 972.21 \\
FairGNN & 3.66 & 5.22 & 4.39 & 793.02 & 1203.24 & 6.65 & 7.18 & 7.54 & 25.98 & 37.97 & 5.44 & 9.54 & 11.08 & 2685.14 & 1203.27 \\
NIFTY & 2.37 & 10.25 & 3.84 & 245.66 & 401.22 & 2.77 & 3.83 & 6.24 & 21.60 & 35.50 & 4.83 & 4.12 & 1.78 & 171.13 & 518.07 \\
FairSIN & 1.87 & 3.84 & 2.84 & 28.89 & 65.36 & 2.50 & 1.22 & 2.73 & 31.01 & 48.53 & 3.89 & 6.19 & 1.82 & 185.50 & 292.76 \\
FairGB & 4.46 & 5.52 & 6.62 & 140.03 & 172.52 & 9.93 & 7.99 & 12.96 & 215.09 & 339.28 & --- & --- & --- & --- & --- \\
Undersampling & 4.89 & 5.78 & 8.35 & 61.54 & 55.93 & 6.58 & 7.15 & 6.18 & 54.42 & 54.42 & 7.13 & 11.09 & 14.38 & 1311.40 & 1296.18 \\
FairDrop & 2.30 & 6.55 & 4.32 & 28.48 & 55.42 & 10.18 & 9.77 & 19.80 & 65.51 & 56.03 & 5.59 & 9.67 & 10.45 & 83.59 & 73.99 \\
BIND & --- & --- & --- & --- & --- & 7.30 & 6.08 & 10.57 & 29.01 & 20.85 & --- & --- & --- & --- & --- \\
\bottomrule
\end{tabular}
}
\end{table*}
\begin{table*}[t]
\centering
\caption{Standard deviation of performance change ratio in early stop condition. `---' indicates no results.}
\label{tab:std_earlystop}
\vspace{-3mm}
\setlength{\tabcolsep}{6pt}
\resizebox{1.0\linewidth}{!}{
\begin{tabular}{lrrrrrrrrrrrrrrr}
\toprule
\multirow{2}{*}{Method} & \multicolumn{5}{c}{Synthetic} & \multicolumn{5}{c}{Social} & \multicolumn{5}{c}{Knowledge} \\
\cmidrule(lr){2-6}\cmidrule(lr){2-6}\cmidrule(lr){7-11}\cmidrule(lr){12-16}
 & ACC & AUC & F1 & $\Delta$SP & $\Delta$Eop & ACC & AUC & F1 & $\Delta$SP & $\Delta$Eop & ACC & AUC & F1 & $\Delta$SP & $\Delta$Eop \\
\midrule
Vanilla & 5.21 & 7.97 & 7.25 & 489.82 & 481.72 & 11.10 & 10.98 & 12.37 & 56.87 & 92.92 & 10.25 & 10.81 & 15.69 & 1283.12 & 1446.43 \\
FairGNN & 5.23 & 5.65 & 5.39 & 1171.77 & 714.32 & 10.02 & 8.91 & 8.00 & 67.04 & 77.67 & 10.21 & 12.60 & 16.90 & 5923.79 & 841.23 \\
NIFTY & 0.80 & 7.54 & 0.89 & 243.34 & 379.51 & 2.66 & 3.20 & 4.46 & 39.97 & 38.19 & 5.90 & 5.20 & 3.34 & 184.17 & 390.56 \\
FairSIN & 0.43 & 1.07 & 0.32 & 89.97 & 115.87 & 2.57 & 1.57 & 3.37 & 80.27 & 63.27 & 6.36 & 5.12 & 5.05 & 787.57 & 514.91 \\
FairGB & 1.12 & 3.38 & 1.70 & 830.62 & 1495.47 & 10.38 & 5.30 & 24.41 & 168.43 & 277.20 & --- & --- & --- & --- & --- \\
Undersampling & 4.64 & 7.22 & 11.64 & 324.63 & 121.71 & 13.72 & 11.07 & 7.07 & 98.86 & 231.18 & 11.05 & 11.76 & 17.17 & 3740.67 & 4054.43 \\
FairDrop & 1.44 & 3.78 & 3.14 & 261.97 & 266.61 & 11.75 & 10.31 & 83.09 & 96.50 & 62.02 & 6.47 & 8.50 & 12.21 & 348.83 & 768.81 \\
BIND & --- & --- & --- & --- & --- & 13.61 & 11.41 & 10.07 & 79.58 & 62.12 & --- & --- & --- & --- & --- \\
\bottomrule
\end{tabular}
}
\end{table*}

\begin{figure*}[t]
    \centering
    \begin{minipage}{0.65\textwidth}
        \centering
        \includegraphics[width=1.0\linewidth]{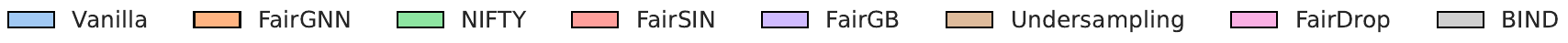}
    \end{minipage}
    \vspace{-1mm}
    \begin{minipage}{0.3\textwidth}
            \centering
    \includegraphics[width=0.87\linewidth]{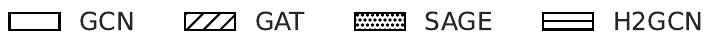}
    \end{minipage}
    \begin{minipage}{1.0\textwidth}
    \centering
    \includegraphics[width=\linewidth]{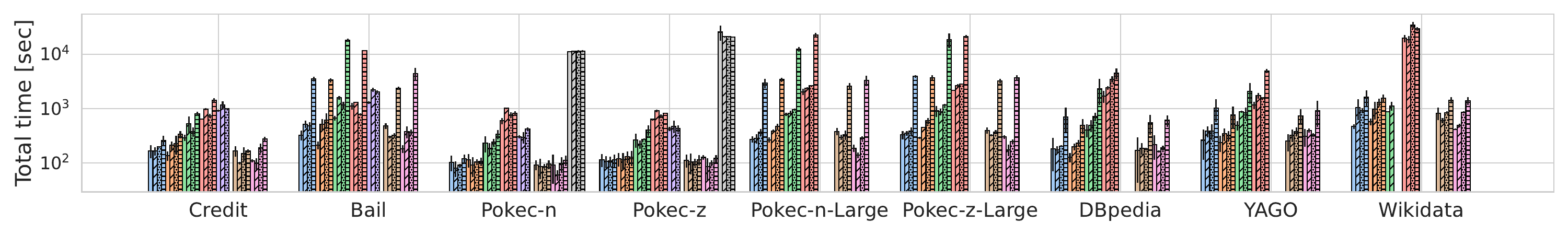}
    \end{minipage}

    \vspace{-5mm}
    \caption{Average total runtime including preprocessing and training}
    \label{fig:totaltime}
    \vspace{-3mm}
\end{figure*}

\begin{table*}
\caption{GPU usage [GB] in GCN and H2GCN. In Wikidata, methods often cannot set the large hidden units as its hyper-parameters, and thus only Wikidata has different trends.}
\label{tab:GPU_GCN_and_H2GCN}
\vspace{-3mm}
\resizebox{1.0\linewidth}{!}{
\begin{tabular}{llccccccccc}
\toprule
Model & Method & Credit & Bail & Pokec-n & Pokec-z & Pokec-n-Large & Pokec-z-Large & DBpedia & YAGO & Wikidata \\
\midrule
\multirow{5}{*}{GCN}&Vanilla & 1.47 $\pm$ 0.03 & 2.60 $\pm$ 0.00 & 0.42 $\pm$ 0.01 & 0.48 $\pm$ 0.02 & 4.45 $\pm$ 0.00 & 5.20 $\pm$ 0.00 & 4.73 $\pm$ 0.09 & 3.41 $\pm$ 0.00 & 33.46 $\pm$ 0.58 \\
&FairGNN & 1.39 $\pm$ 0.04 & 2.00 $\pm$ 0.00 & 0.43 $\pm$ 0.01 & 0.49 $\pm$ 0.00 & 4.15 $\pm$ 0.15 & 4.90 $\pm$ 0.14 & 4.28 $\pm$ 0.11 & 3.10 $\pm$ 0.15 & 34.05 $\pm$ 1.31 \\
&NIFTY & 1.75 $\pm$ 0.02 & 2.60 $\pm$ 0.00 & 0.55 $\pm$ 0.01 & 0.65 $\pm$ 0.02 & 5.20 $\pm$ 0.00 & 5.96 $\pm$ 0.29 & 8.09 $\pm$ 0.38 & 5.21 $\pm$ 0.31 & 31.60 $\pm$ 0.00 \\
&FairSIN & 1.32 $\pm$ 0.00 & 2.47 $\pm$ 0.00 & 0.39 $\pm$ 0.00 & 0.45 $\pm$ 0.00 & 4.60 $\pm$ 0.00 & 5.09 $\pm$ 0.00 & 3.68 $\pm$ 0.00 & 2.83 $\pm$ 0.00 & 33.85 $\pm$ 1.41 \\
&FairGB & 19.77 $\pm$ 0.00 & 10.38 $\pm$ 0.00 & 1.06 $\pm$ 0.00 & 1.50 $\pm$ 0.00 & - & - & - & - & - \\\hline

\multirow{5}{*}{H2GCN}&Vanilla & 4.12 $\pm$ 0.33 & 8.28 $\pm$ 0.09 & 1.05 $\pm$ 0.06 & 1.22 $\pm$ 0.02 & 13.61 $\pm$ 0.11 & 15.83 $\pm$ 0.96 & 21.26 $\pm$ 2.83 & 18.42 $\pm$ 1.62 & 31.20 $\pm$ 0.43 \\
&FairGNN & 3.90 $\pm$ 0.20 & 9.22 $\pm$ 0.06 & 1.05 $\pm$ 0.02 & 1.21 $\pm$ 0.04 & 14.54 $\pm$ 0.44 & 16.58 $\pm$ 0.14 & 19.89 $\pm$ 0.85 & 15.90 $\pm$ 0.40 & 25.14 $\pm$ 4.99 \\
&NIFTY & 4.91 $\pm$ 0.06 & 9.15 $\pm$ 0.12 & 1.35 $\pm$ 0.08 & 1.72 $\pm$ 0.05 & 16.77 $\pm$ 0.44 & 18.66 $\pm$ 0.78 & 30.23 $\pm$ 0.89 & 22.34 $\pm$ 0.95 & - \\
&FairSIN & 1.99 $\pm$ 0.00 & 4.46 $\pm$ 0.00 & 0.60 $\pm$ 0.00 & 0.70 $\pm$ 0.00 & 6.15 $\pm$ 0.10 & 6.43 $\pm$ 0.00 & 9.44 $\pm$ 0.00 & 7.83 $\pm$ 0.00 & 42.23 $\pm$ 0.23 \\
&FairGB & - & - & - & - & - & - & - & - & - \\

\bottomrule
\end{tabular}
}
\vspace{-3mm}
\end{table*}

\noindent
{\bf Q5. Learning scalability}.
We evaluate the scalability in terms of learning time and GPU memory usage.
Figure~\ref{fig:totaltime} shows the total runtime that includes preprocessing and training time.
Table~\ref{tab:GPU_GCN_and_H2GCN} shows the GPU memory usage in GCN and H2GCN.

From Figure~\ref{fig:totaltime}, we can see that BIND takes significantly longer than other methods; that is why we evaluate BIND only in Pokec-n and Pokec-z.
FairSIN takes a longer time than other in-processing methods because FairSIN has a graph modification process, which may cause a long learning time. Comparing backbone models, H2GCN takes much longer than other models on all methods, so model choice must be considered for learning time.
The learning efficiency is often ignored in fairness-aware GNN studies, while we need efficient and fair learning algorithms in large-scale graphs.

From Table~\ref{tab:GPU_GCN_and_H2GCN}, FairGB and NIFTY use larger GPU memory among them, while FairSIN uses smaller GPU memory than the others.
It is well-known that H2GCN utilizes larger GPU memory than GCN; however, the memory usage of fairness-aware methods on H2GCN has not been well studied.
Since each method has different GPU memory usage, we need to consider the method as well as the models.
This scalability analysis demonstrates how existing methods can be applied to large-scale datasets.

\section{Conclusion, Limitation, and Broader Impact}
\label{sec:conclusion}

We introduced new graph datasets generated from knowledge graphs and conducted benchmarking studies for fairness-aware GNN methods.
Our new graphs revealed distinct trends in trade-offs between prediction accuracy and fairness metrics for fairness-aware GNNs.
Our benchmarks examined the impact of fairness-aware GNN methods, GNN backbone models, and early stopping conditions. We demonstrated that not only fairness-aware GNN methods, but also GNN backbones and early stopping conditions, have a significant impact on fairness performance.

\noindent
{\bf Impact on fairness-aware graph learning}.
Benchmarking studies using large public datasets often advance machine learning research.
Similar to other studies, our new datasets and benchmarking facilitate the development of fairness-aware GNNs.

\noindent
{\bf Limitation}.
This study has several limitations.
First, we only focus on a node classification task with binary labels and binary-sensitive attributes. Although this is the most common task, it is beneficial to perform more complex tasks, such as classification with multiple labels and multiple sensitive attributes.
Second, the number of baselines is limited; therefore, we plan to expand our framework to include additional baselines. 
Third, we evaluate group fairness using statistical parity and equal opportunity. 
We plan to assess other types of fairness metrics, such as counterfactual fairness.
Finally, we do not propose new fairness-aware methods for knowledge graphs. We will develop new methods that are suitable for knowledge graphs. 

\noindent
{\bf Negative societal impact}.
Our work may have a negative societal impact, particularly in the context of graph neural networks.
Our study provided biased datasets related to gender and nationality.
If users use these biased datasets without caution, the trained models may produce biased predictions. On the other hand, our datasets can be used to raise awareness and check/mitigate the bias in models.


\bibliography{graphassociationrule}
\bibliographystyle{plain}




\section{Detailed results}
\label{sec:detailedresults}

\begin{figure*}[t!]
    \centering
    \begin{minipage}{\textwidth}
        \centering
        \includegraphics[width=0.9\linewidth]{figures/legend_only.pdf}
    \end{minipage}
    \begin{minipage}{0.3\textwidth}
        \includegraphics[width=0.8\linewidth]{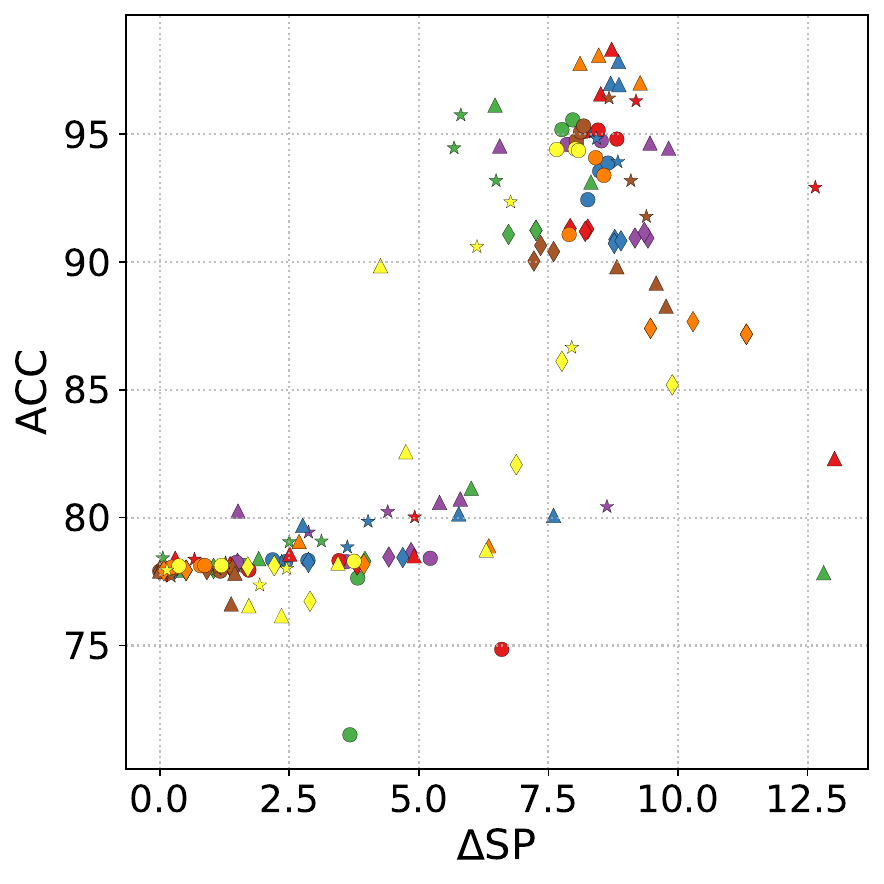}
        \subcaption{$\Delta$SP ($\downarrow$) and ACC ($\uparrow$) in synthetic graphs}
    \end{minipage}
        \begin{minipage}{0.3\textwidth}
        \includegraphics[width=0.8\linewidth]{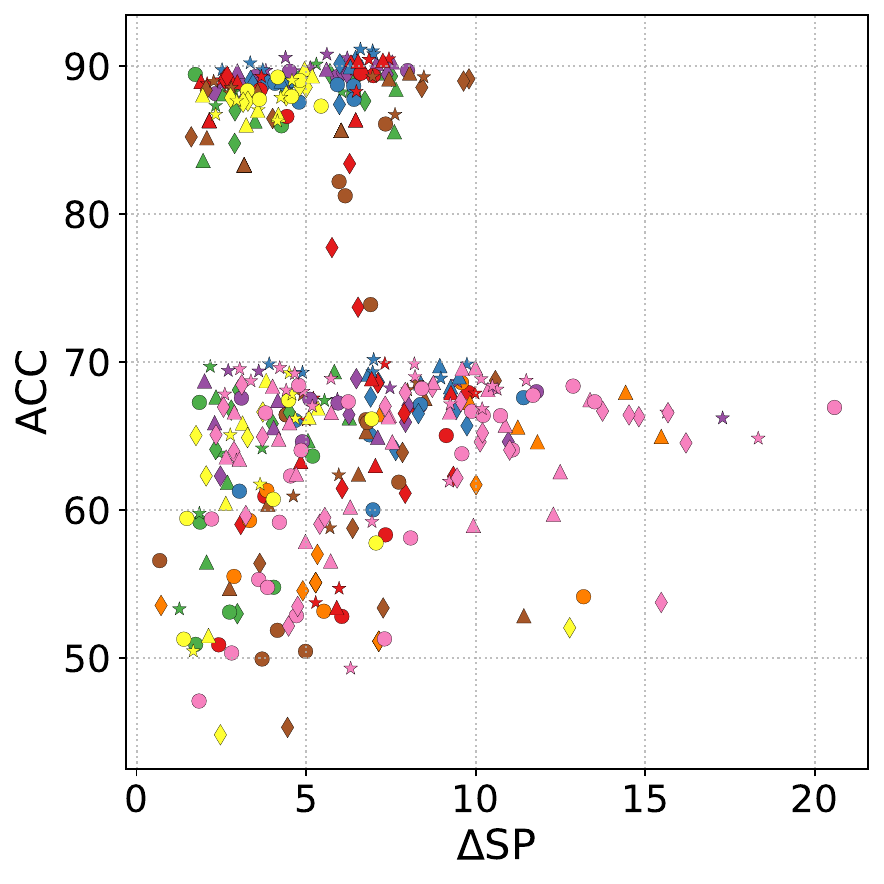}
        \subcaption{$\Delta$SP ($\downarrow$) and ACC ($\uparrow$) in social graphs}
    \end{minipage}
        \begin{minipage}{0.3\textwidth}
        \includegraphics[width=0.8\linewidth]{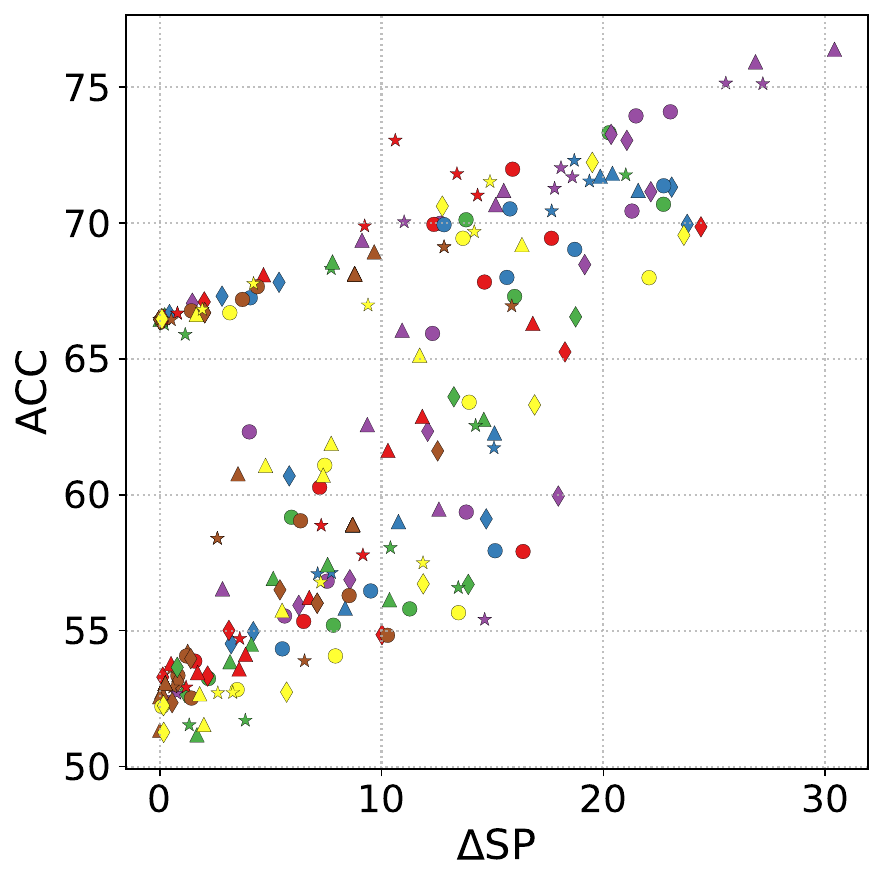}
        \subcaption{$\Delta$SP ($\downarrow$) and ACC ($\uparrow$) in knowledge graphs}
    \end{minipage}

   \begin{minipage}{0.3\textwidth}
        \includegraphics[width=0.8\linewidth]{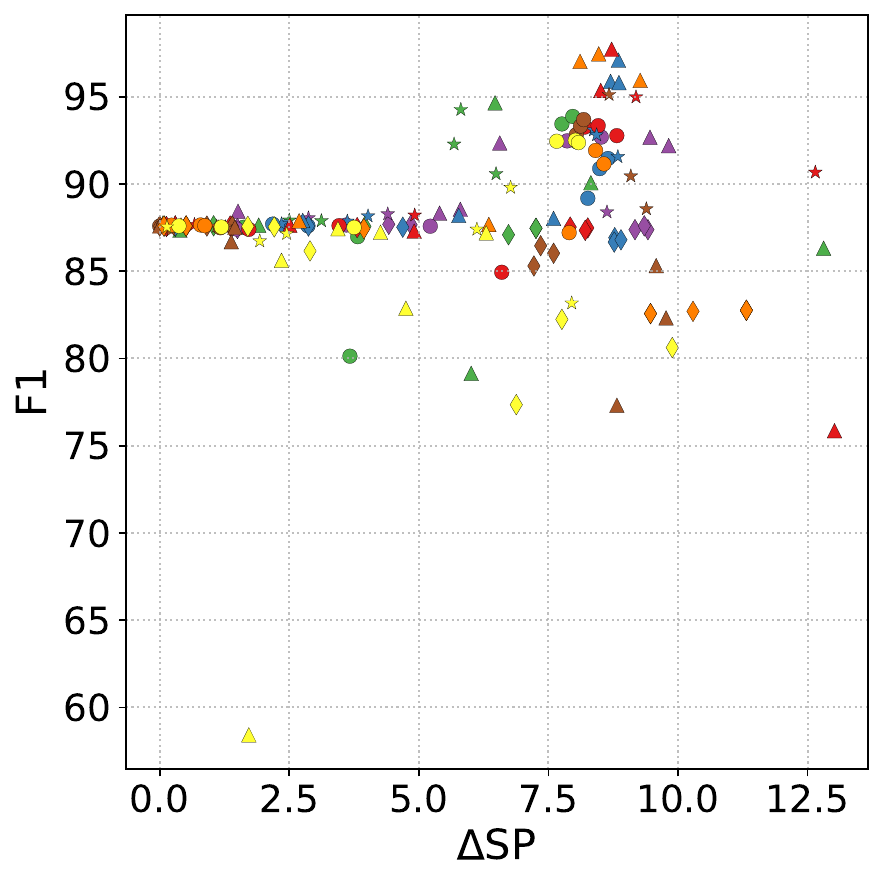}
        \subcaption{$\Delta$SP ($\downarrow$) and F1 ($\uparrow$) in Synthetic graphs}
    \end{minipage}
        \begin{minipage}{0.3\textwidth}
        \includegraphics[width=0.8\linewidth]{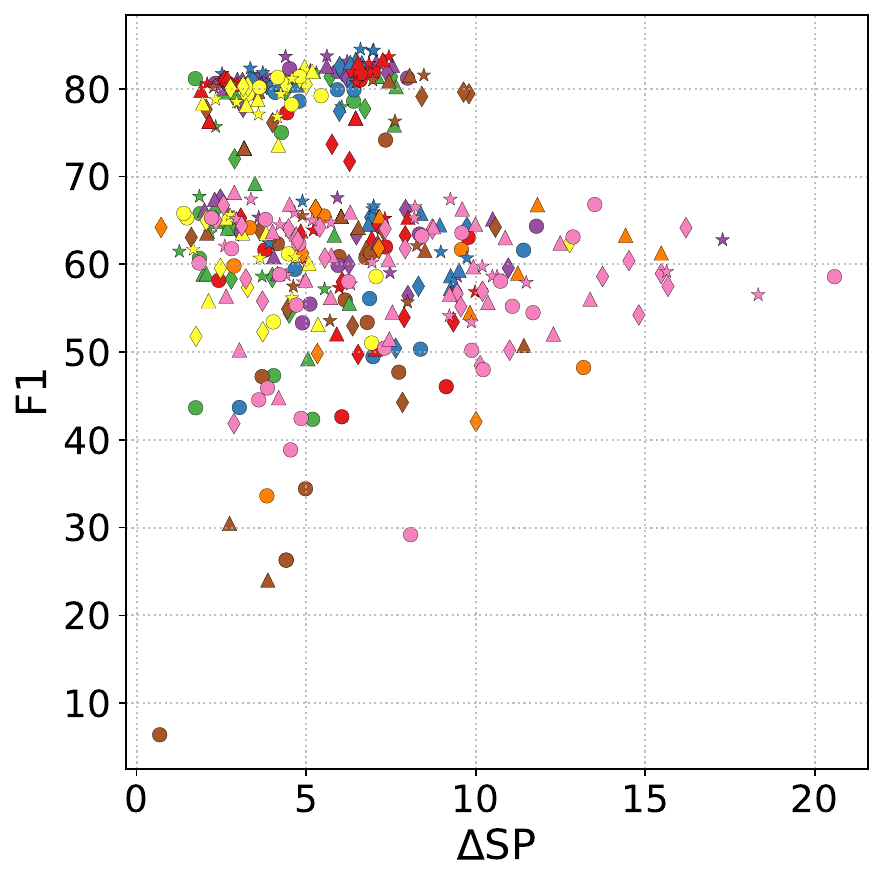}
        \subcaption{$\Delta$SP ($\downarrow$) and F1 ($\uparrow$) in social graphs}
    \end{minipage}
        \begin{minipage}{0.3\textwidth}
        \includegraphics[width=0.8\linewidth]{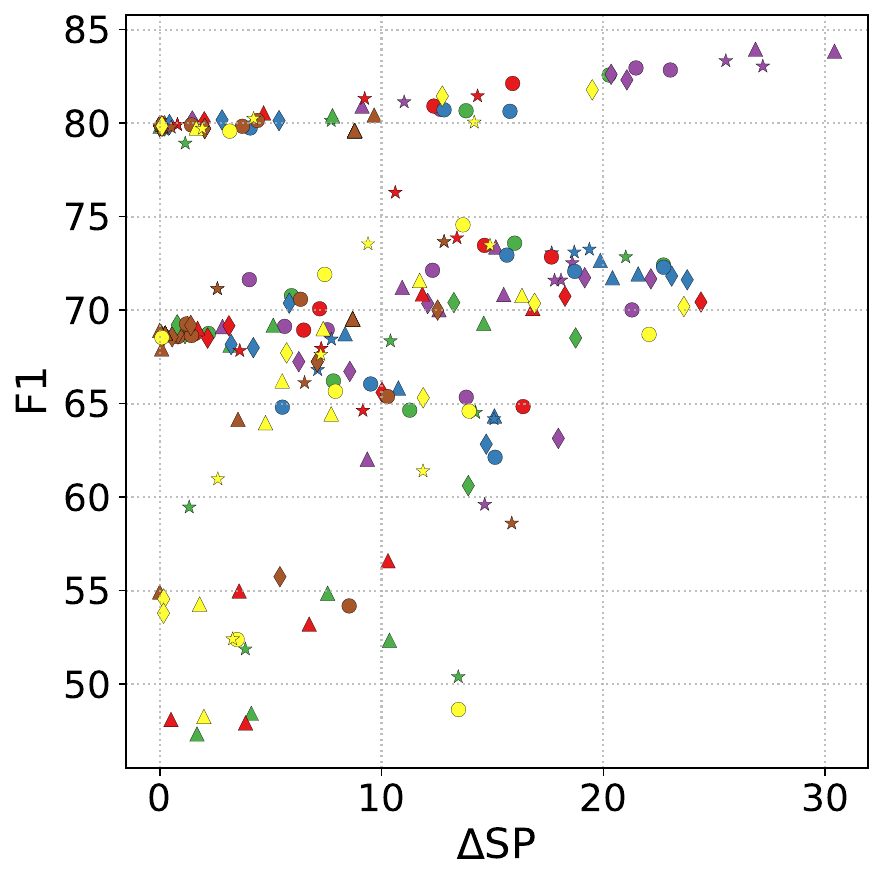}
        \subcaption{$\Delta$SP ($\downarrow$) and F1 ($\uparrow$) in knowledge graphs}
    \end{minipage}

        \begin{minipage}{0.3\textwidth}
        \includegraphics[width=0.8\linewidth]{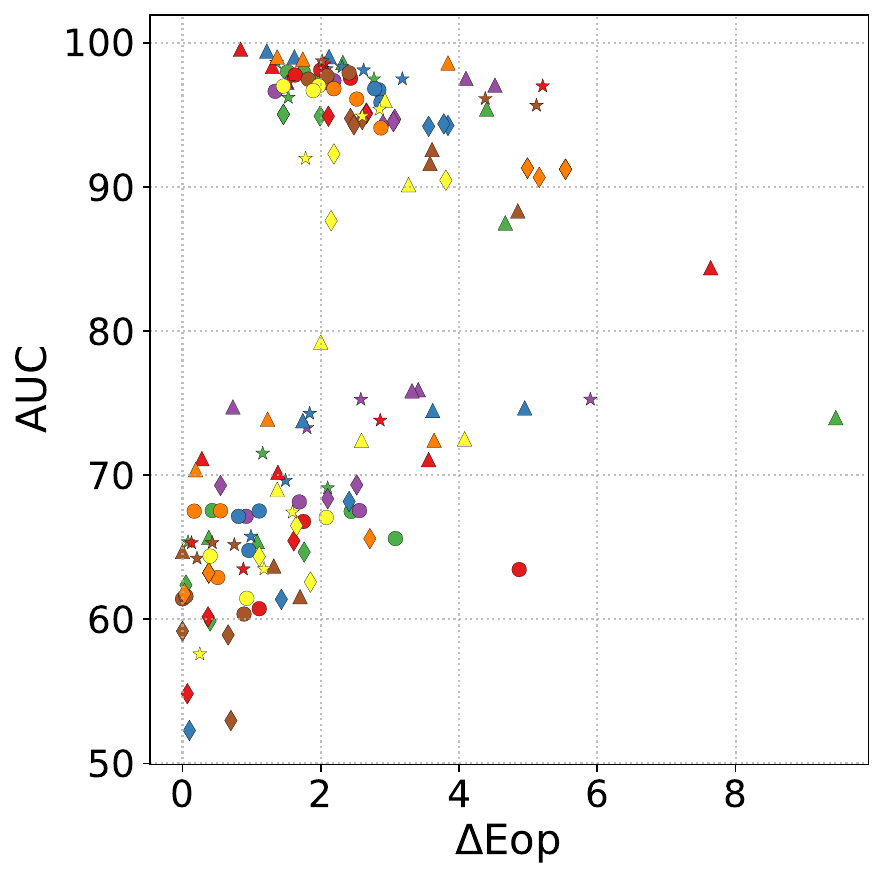}
        \subcaption{$\Delta$Eop ($\downarrow$) and AUC ($\uparrow$) in synthetic graphs}
    \end{minipage}
        \begin{minipage}{0.3\textwidth}
        \includegraphics[width=0.8\linewidth]{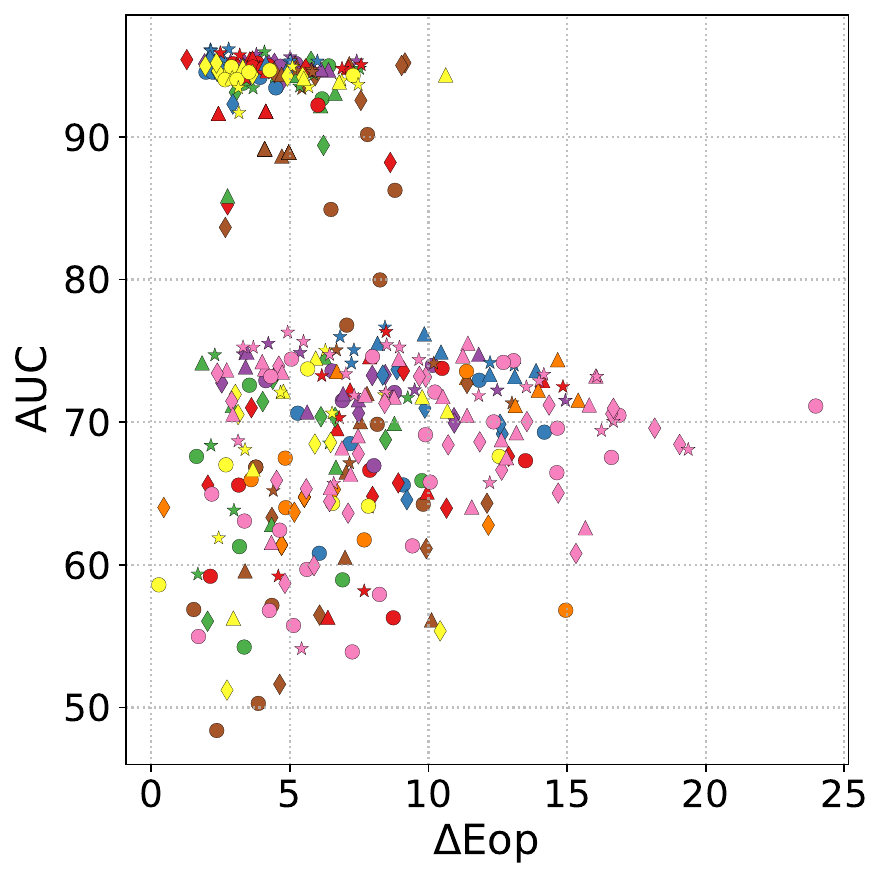}
        \subcaption{$\Delta$Eop ($\downarrow$) and AUC ($\uparrow$) in social graphs}
    \end{minipage}
        \begin{minipage}{0.3\textwidth}
        \includegraphics[width=0.8\linewidth]{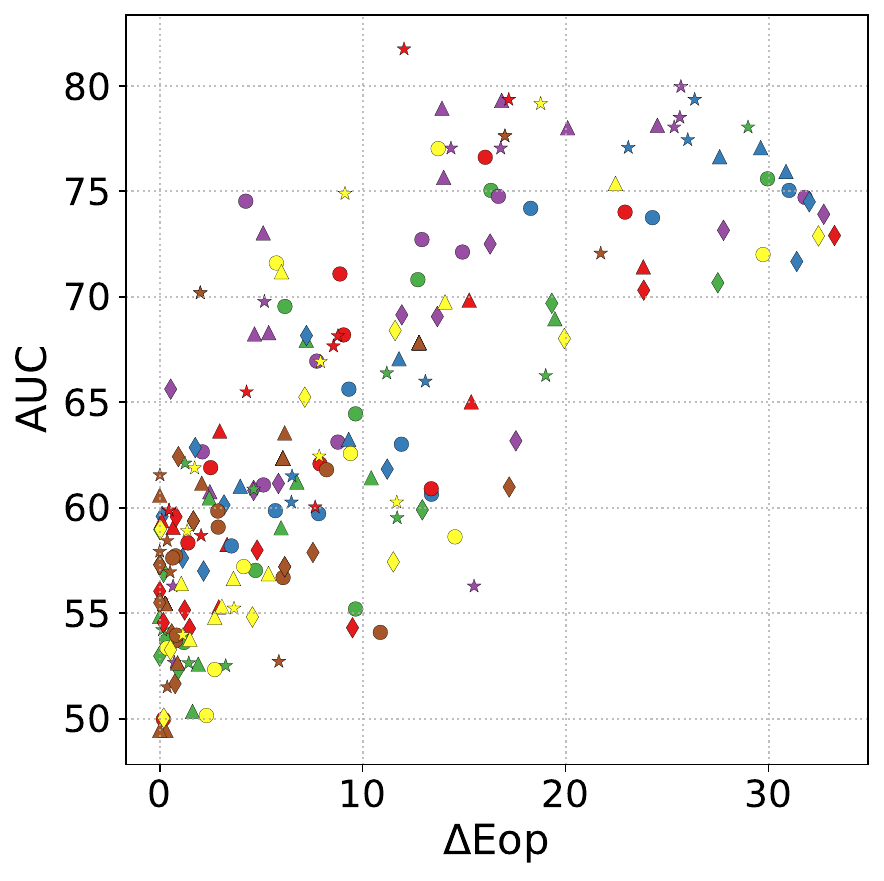}
        \subcaption{$\Delta$Eop ($\downarrow$) and AUC ($\uparrow$) in knowledge graphs}
    \end{minipage}

   \begin{minipage}{0.3\textwidth}
        \includegraphics[width=0.8\linewidth]{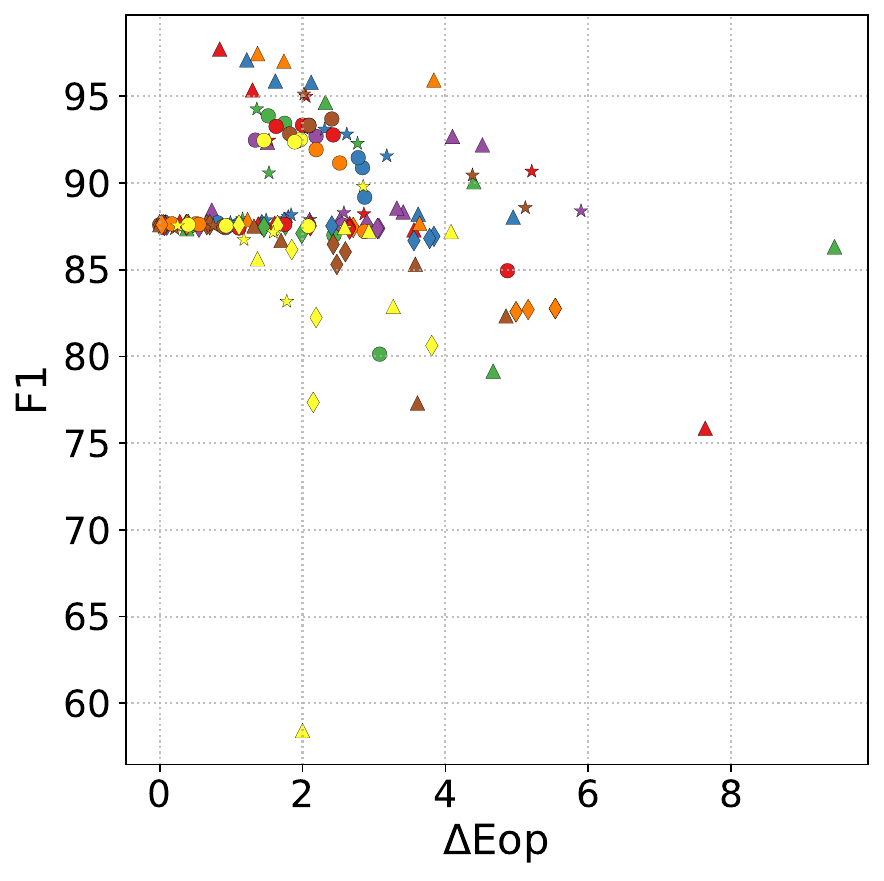}
        \subcaption{$\Delta$Eop ($\downarrow$) and F1 ($\uparrow$) in synthetic graphs}
    \end{minipage}
        \begin{minipage}{0.3\textwidth}
        \includegraphics[width=0.8\linewidth]{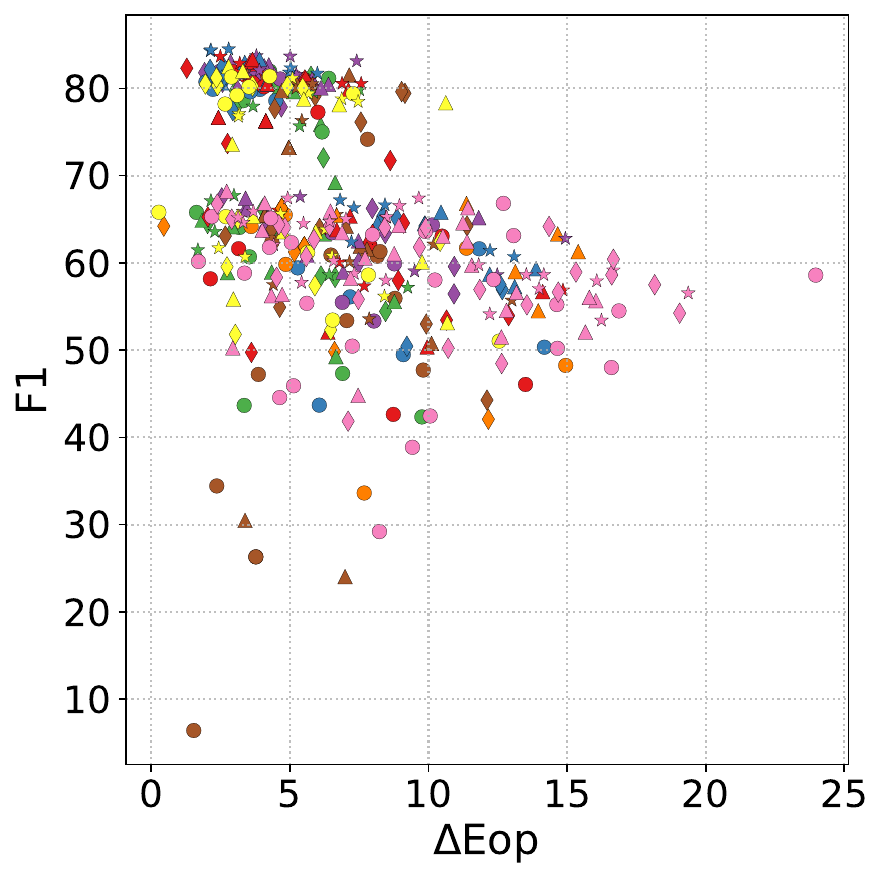}
        \subcaption{$\Delta$Eop ($\downarrow$) and F1 ($\uparrow$) in social graphs}
    \end{minipage}
        \begin{minipage}{0.3\textwidth}
        \includegraphics[width=0.8\linewidth]{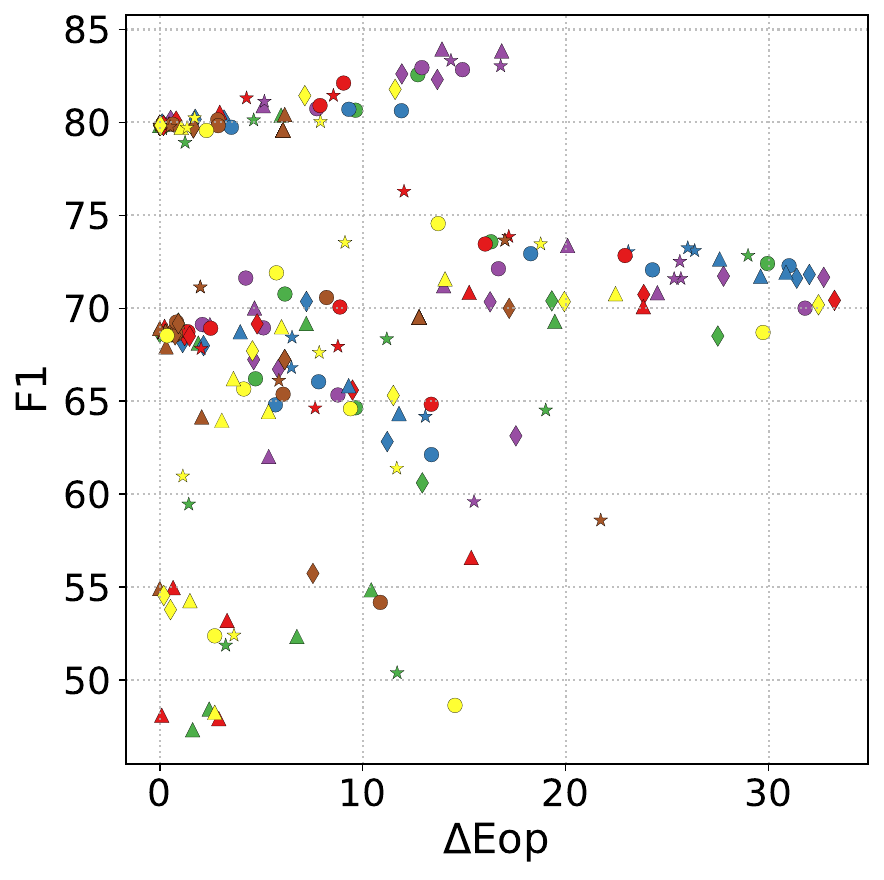}
        \subcaption{$\Delta$Eop ($\downarrow$) and F1 ($\uparrow$) in knowledge graphs}
    \end{minipage}

    \caption{(Additional) The overall trends of trade-offs in dataset types.}
    \label{fig:result_tradeoff_appendix}
\end{figure*}

Figure~\ref{fig:result_tradeoff_appendix} shows the trade-offs between other prediction accuracy and fairness metrics, which have different trends. Table~\ref{tab:tradeoff_pearson_appendix} shows the Pearson correlations. Knowledge graphs have clearer tradeoffs compared to other graph types in ACC-$\Delta$SP and AUC-$\Delta$Eop. Interestingly, F1 does not have strong tradeoffs compared with ACC and AUC, and many graphs including DBpedia have negative values, indicating F1 and fairness can be improved simultaneously.

Table~\ref{table:result_H2GCN_Alpha} shows the performance in each method with H2GCN backbone and Mixed early stop condition.
We skip other results due to the page limitations.
Compared to Table~\ref{table:result_GCN_ACC}, fairness metrics becomes much better, for example, $\Delta$Eop of FairDrop in Credit changes from 0.70 to 0.21, indicating that the early stopping condition including fairness metrics improve fairness.

\section{Hyper parameters}
\label{sec:hyperparameters}

We select hyper-parameters for the search space according to their papers or codebases. We set the number of layers in GNNs to two.
If they need to specify values instead of ranges, we set fixed values.
More concretely, see our codebases.

\smallskip\noindent
\textbf{Vanilla and Undersampling.}
\begin{itemize}
    \item Hidden channel: $[16, 64, 128, 256]$
    \item Proj Hidden channel: $[16, 64, 128, 256]$
    \item Learning rate: $[1e^{-2}, 1e^{-3}, 1^{-4}, 1e^{-5}]$
    \item Weight Decay: $[1e^{-2}, 5e^{-2}, 1e^{-3}, 2e^{-3}, 1e^{-4}, 1e^{-5}]$
    \item Sim Coeff: $[0.3, 0.5, 0.7]$
    \item Epochs: $[2000]$
    \item Early Stopping: $[20]$
     \item Dropout: $[0.5]$
\end{itemize}

\textbf{FairGNN. }
\begin{itemize}
    \item Hidden channel: $[16, 64, 128, 256]$
    \item Proj Hidden channel: $[16, 32, 64, 128, 256]$
    \item Acc: $[0.2, 0.3, 0.4, 0.5, 0.6, 0.68, 0.688, 0.69, 0.7, 0.8, 0.9]$
    \item Alpha: $[1, 2, 3, 4, 5, 6, 7, 10, 20, 40, 50, 100]$
    \item Beta: $[1, 0.1, 0.01, 0.001, 0.0001]$
    \item Learning rate: $[1e^{-2}, 1e^{-3}, 1^{-4}, 1e^{-5}]$
    \item Weight Decay: $[1e^{-2}, 1e^{-3}, 1^{-4}, 1e^{-5}]$
    \item Sim Coeff: $[0.3, 0.5, 0.6, 0.7]$
    \item Epochs: $[2000]$
    \item Early Stopping: $[20]$
     \item Dropout: $[0.5]$
\end{itemize}

\textbf{NIFTY. }
\begin{itemize}
    \item Hidden channel: $[16, 32, 64, 128, 256]$
    \item Proj Hidden channel: $[16, 32, 64, 128, 256]$
    \item Drop Edge Rate1: $[0.1, 0.01, 0.001, 0.0001]$
    \item Drop Edge Rate2: $[0.1, 0.01, 0.001, 0.0001]$
    \item Drop Feature Rate1: $[0.1, 0.01, 0.001, 0.0001]$
    \item Drop Feature Rate2: $[0.1, 0.01, 0.001, 0.0001]$
    \item Learning rate: $[1e^{-2}, 1e^{-3}, 1^{-4}, 1e^{-5}]$
    \item Weight Decay: $[1e^{-2}, 1e^{-3}, 1^{-4}, 1e^{-5}, 1e^{-6}]$
    \item Sim Coeff: $[0.3, 0.4, 0.5, 0.6, 0.7]$
    \item Epochs: $[2000]$
    \item Early Stopping: $[20]$
     \item Dropout: $[0.5]$
\end{itemize}

\begin{table}
\caption{(Additional) Pearson correlation between performance and fairness metrics. Values in parentheses denote $p$-values.}
\label{tab:tradeoff_pearson_appendix}
\vspace{-3mm}
\resizebox{1.0\linewidth}{!}{
\begin{tabular}{lcccc}
\toprule
Dataset & ACC--$\Delta$SP & AUC--$\Delta$Eop & F1--$\Delta$SP & F1--$\Delta$Eop \\
\midrule
Credit & 0.23 (0.032) & 0.57 (0.000) & -0.08 (0.454) & -0.18 (0.103) \\
Bail & 0.11 (0.306) & -0.54 (0.000) & 0.13 (0.235) & -0.48 (0.000) \\
Pokec-n & 0.38 (0.000) & 0.54 (0.000) & 0.22 (0.033) & 0.14 (0.165) \\
Pokec-z & 0.30 (0.003) & 0.36 (0.000) & -0.16 (0.115) & -0.20 (0.048) \\
Pokec-n-Large & -0.33 (0.002) & 0.04 (0.700) & -0.22 (0.043) & 0.01 (0.915) \\
Pokec-z-Large & 0.05 (0.632) & -0.19 (0.082) & 0.07 (0.536) & -0.22 (0.049) \\
DBpedia & 0.84 (0.000) & 0.57 (0.000) & -0.25 (0.024) & -0.30 (0.006) \\
YAGO & 0.88 (0.000) & 0.64 (0.000) & 0.34 (0.002) & 0.29 (0.008) \\
Wikidata & 0.67 (0.000) & 0.83 (0.000) & 0.26 (0.024) & 0.24 (0.034) \\
\bottomrule
\end{tabular}
}
\end{table}
\begin{table*}[!t]
\caption{Prediction accuracy and fairness metrics on H2GCN backbone and Mixed early stopping. Bold and underlined values indicate the best and second-best methods, respectively. OOM, DNF, and `---' indicate ``out-of-memory errors'', ``did not finish within 24 hours'', and 'did not work due to incompatibility of method and model', respectively.}\label{table:result_H2GCN_Alpha}
\begin{center}
\resizebox{1.0\linewidth}{!}{

\begin{tabular}{ c | c | c c c c c c c c c } \hline
 Metric  & Method & Credit & Bail & Pokec-n & Pokec-z & Pokec-n-Large & Pokec-z-Large & YAGO & DBpedia & Wikidata \\ \hline\hline
\multirow{8}{*}{ACC}&Vanilla & 77.97±0.00 & 94.67±0.22 & 66.08±0.41 & 67.56±0.04 & 88.81±0.02 & 88.28±0.16 & 58.87±0.24 & 52.92±0.01 & 66.67±0.00 \\
&FairGNN & \underline{78.42±0.01} & \underline{95.74±0.01} & 64.19±0.32 & 63.84±0.64 & 88.29±0.02 & 88.26±0.21 & 58.05±0.11 & 51.53±0.01 & 65.89±0.04 \\
&NIFTY & 78.28±0.00 & 93.91±0.05 & \textbf{69.88±0.01} & \underline{69.30±0.00} & \underline{89.75±0.01} & \textbf{90.82±0.00} & \underline{70.43±0.01} & \textbf{57.09±0.00} & OOM \\
&FairSIN & \textbf{79.43±0.01} & 95.06±0.03 & \underline{68.26±0.01} & \textbf{69.43±0.00} & \textbf{89.89±0.00} & \underline{90.55±0.01} & \textbf{71.26±0.00} & 52.54±0.00 & \textbf{70.03±0.02} \\
&FairGB & ---& ---& ---& ---& ---& ---& ---& ---& ---\\
&Undersampling & 77.94±0.00 & 86.65±1.09 & 66.18±0.03 & 61.75±0.41 & 86.30±0.03 & 89.03±0.01 & 56.78±0.57 & 52.72±0.06 & \underline{66.81±0.01} \\
&FairDrop & 77.71±0.00 & \textbf{96.40±0.04} & 60.94±0.68 & 67.79±0.00 & 88.54±0.03 & 86.73±0.09 & 58.39±0.29 & \underline{53.60±0.02} & 66.47±0.00 \\
&BIND & DNF & DNF & 68.16±0.00 & 67.02±0.04 & DNF & DNF & DNF & DNF & DNF \\
\hline
\multirow{8}{*}{F1}&Vanilla & 63.48±0.94 & 97.24±0.14 & 70.31±0.51 & 73.26±0.17 & 94.79±0.00 & 94.05±0.09 & 68.15±0.45 & 55.58±0.16 & 59.88±0.45 \\
&FairGNN & 65.36±0.68 & \underline{98.63±0.00} & 68.53±0.59 & 68.37±1.73 & 94.79±0.00 & 93.42±0.23 & 66.38±0.21 & 52.64±0.03 & \underline{62.11±0.23} \\
&NIFTY & \underline{65.75±0.06} & 97.49±0.02 & \textbf{74.13±0.00} & \textbf{75.99±0.01} & \textbf{95.30±0.00} & \textbf{96.03±0.00} & \underline{77.07±0.01} & \textbf{60.26±0.04} & OOM \\
&FairSIN & \textbf{73.28±0.22} & 98.21±0.01 & 72.26±0.00 & 74.77±0.01 & \underline{94.84±0.00} & \underline{95.31±0.01} & \textbf{78.04±0.02} & 52.66±0.06 & \textbf{69.77±0.12} \\
&FairGB & ---& ---& ---& ---& ---& ---& ---& ---& ---\\
&Undersampling & 57.58±1.04 & 91.97±0.57 & 70.63±0.07 & 68.07±0.63 & 93.44±0.02 & 94.92±0.01 & 62.44±1.06 & 55.23±0.19 & 58.90±0.12 \\
&FairDrop & 64.23±0.68 & \textbf{98.73±0.02} & 65.20±1.00 & \underline{75.05±0.00} & 94.54±0.01 & 93.38±0.04 & 70.18±0.47 & \underline{58.44±0.14} & 61.55±0.23 \\
&BIND & DNF & DNF & \underline{73.34±0.01} & 73.38±0.12 & DNF & DNF & DNF & DNF & DNF \\
\hline
\multirow{8}{*}{AUC}&Vanilla & 87.53±0.00 & 92.46±0.48 & 60.05±0.17 & 63.88±0.38 & 80.56±0.02 & 80.71±0.24 & 67.95±0.33 & \underline{68.66±0.00} & \underline{79.92±0.00} \\
&FairGNN & \underline{87.76±0.00} & \underline{94.26±0.01} & 58.63±0.18 & \underline{67.11±0.08} & 79.84±0.04 & 77.94±1.10 & 68.35±0.02 & 59.46±1.09 & 78.92±0.05 \\
&NIFTY & 87.71±0.00 & 91.56±0.09 & \textbf{62.40±0.01} & \textbf{67.19±0.02} & \underline{81.72±0.01} & \textbf{84.38±0.01} & \textbf{73.05±0.00} & 66.81±0.04 & OOM \\
&FairSIN & \textbf{88.05±0.00} & 93.15±0.08 & 59.05±0.19 & 65.86±0.07 & \textbf{82.06±0.01} & \underline{83.13±0.03} & \underline{71.59±0.00} & 68.58±0.00 & \textbf{81.13±0.01} \\
&FairGB & ---& ---& ---& ---& ---& ---& ---& ---& ---\\
&Undersampling & 87.54±0.00 & 83.17±1.51 & \underline{60.79±0.08} & 60.69±0.41 & 76.79±0.03 & 81.14±0.04 & 67.62±0.26 & 52.41±4.38 & 79.71±0.00 \\
&FairDrop & 87.36±0.00 & \textbf{95.12±0.07} & 57.50±0.56 & 65.53±0.08 & 80.52±0.04 & 76.31±0.44 & 71.14±0.04 & \textbf{68.72±0.00} & 79.85±0.00 \\
&BIND & DNF & DNF & 58.64±0.02 & 65.01±0.10 & DNF & DNF & DNF & DNF & DNF \\
\hline
\multirow{8}{*}{$\Delta$SP}&Vanilla & 1.14±0.03 & 8.03±0.05 & 5.00±0.08 & 5.21±0.07 & \textbf{2.08±0.02} & 6.48±0.05 & 7.28±0.17 & 1.19±0.02 & \underline{0.80±0.02} \\
&FairGNN & \textbf{0.06±0.00} & \textbf{5.81±0.10} & \textbf{3.70±0.04} & \textbf{2.44±0.05} & 2.88±0.02 & \underline{6.05±0.01} & 10.40±0.54 & 1.33±0.01 & 1.15±0.02 \\
&NIFTY & 2.35±0.03 & 8.84±0.05 & \underline{3.91±0.12} & 4.89±0.17 & \underline{2.52±0.02} & 6.99±0.07 & 17.67±0.10 & 7.12±0.04 & OOM \\
&FairSIN & 2.87±0.04 & 8.44±0.10 & 7.47±0.13 & \underline{2.70±0.04} & 5.12±0.10 & 6.21±0.11 & 17.80±0.13 & \textbf{0.23±0.00} & 11.02±0.14 \\
&FairGB & ---& ---& ---& ---& ---& ---& ---& ---& ---\\
&Undersampling & \underline{0.14±0.00} & \underline{7.95±0.03} & 4.68±0.03 & 3.64±0.20 & 4.15±0.03 & \textbf{4.58±0.01} & \underline{7.26±0.53} & 3.29±0.15 & 1.92±0.14 \\
&FairDrop & 0.22±0.00 & 8.67±0.01 & 4.62±0.02 & 4.89±0.08 & 2.79±0.01 & 7.62±0.41 & \textbf{2.60±0.06} & \underline{0.53±0.00} & \textbf{0.07±0.00} \\
&BIND & DNF & DNF & 10.63±0.04 & 5.18±0.07 & DNF & DNF & DNF & DNF & DNF \\
\hline
\multirow{8}{*}{$\Delta$Eop}&Vanilla & 0.88±0.03 & \underline{1.53±0.01} & 6.78±0.23 & 6.15±0.15 & 6.88±0.08 & \underline{3.43±0.02} & 8.78±0.20 & \textbf{0.08±0.00} & \underline{0.46±0.01} \\
&FairGNN & \textbf{0.08±0.00} & \textbf{1.36±0.02} & \underline{6.42±0.10} & \textbf{2.15±0.02} & 7.51±0.10 & 3.67±0.03 & 11.19±0.94 & 1.43±0.03 & 1.25±0.01 \\
&NIFTY & 0.99±0.01 & 3.18±0.05 & 7.22±0.17 & 6.81±0.20 & 5.99±0.05 & \textbf{2.15±0.04} & 23.09±0.21 & 6.49±0.03 & OOM \\
&FairSIN & 1.80±0.01 & 2.02±0.01 & 9.50±0.31 & \underline{3.31±0.04} & \underline{3.90±0.09} & 7.41±0.22 & 25.35±0.25 & 0.72±0.02 & 5.16±0.10 \\
&FairGB & ---& ---& ---& ---& ---& ---& ---& ---& ---\\
&Undersampling & 0.25±0.00 & 1.78±0.05 & 6.52±0.11 & 3.38±0.12 & \textbf{3.14±0.02} & 5.09±0.06 & \underline{7.86±0.72} & 3.66±0.15 & 1.35±0.06 \\
&FairDrop & \underline{0.21±0.00} & 2.02±0.02 & \textbf{4.40±0.10} & 6.68±0.07 & 5.81±0.16 & 5.43±0.55 & \textbf{2.00±0.07} & \underline{0.37±0.00} & \textbf{0.01±0.00} \\
&BIND & DNF & DNF & 14.16±0.24 & 7.02±0.10 & DNF & DNF & DNF & DNF & DNF \\
\hline
\end{tabular}
}
\end{center}
\end{table*}

\textbf{FairSIN. }
\begin{itemize}
    \item Hidden channel: $[16, 32, 64, 128, 256]$
    \item C Learning rate: $[1e^{-2}, 1e^{-3}, 1^{-4}, 1e^{-5}]$
    \item E Learning rate: $[1e^{-2}, 1e^{-3}, 1^{-4}, 1e^{-5}]$
    \item M Learning rate: $[1e^{-2}, 1e^{-3}, 1^{-4}, 1e^{-5}]$
    \item D Learning rate: $[1e^{-2}, 1e^{-3}, 1^{-4}, 1e^{-5}]$
    \item Delta: $[0.5, 1, 5]$
    \item Alpha: $[0.5, 1, 3]$
    \item Weight Decay: $[0.001, 0.0001, 0]$
    \item Epochs: $[50]$
    \item C Epochs: $[10]$
    \item D Epochs: $[10]$
    \item M Epochs: $[20]$
    \item Early Stopping: $[20]$
    \item Dropout: $[0.2, 0.5, 0.8]$
\end{itemize}

\textbf{FairGB. }
\begin{itemize}
    \item Hidden channel: $[16, 32, 64, 128, 256]$
    \item E Learning rate: $[1e^{-2}, 1e^{-3}]$
    \item C Learning rate: $[1e^{-2}, 1e^{-3}]$
    \item Weight Decay: $[0.0001, 0.00001]$
    \item Warmup: $[5]$
    \item eta: $[0.5]$
\end{itemize}

\textbf{FairDrop. }
\begin{itemize}
    \item Hidden channel: $[16, 64, 128, 256]$
    \item GNN Hidden channel: $[16, 64, 128, 256]$
    \item Learning rate: $[1e^{-2}, 1e^{-3}, 1^{-4}, 1e^{-5}]$
    \item Weight Decay: $[1e^{-2}, 5e^{-2}, 1e^{-3}, 2e^{-3}, 1e^{-4}, 1e^{-5}]$
    \item Delta: $[0.1, 0.2, 0.3, 0.4]$
\end{itemize}

\textbf{BIND. }
BIND takes a long time for preprocessing; therefore, we used the parameters provided in their paper and codebase. 
\begin{itemize}
    \item bind del rate: $[10]$
    \item bind fastmode: False
    \item bind seed: $[10]$
    \item bind epochs: $[1000]$
    \item bind lr: $[0.001]$
    \item bind weight decay: $[0.0001]$
    \item bind hidden: $[16]$
    \item bind dropout: $[0.5]$
    \item bind helpfulness collection: 0   
\end{itemize}

\end{document}